# *Delving* Into the Psychology of Machines: Exploring the Structure of Self-Regulated Learning via LLM-Generated Survey Responses

Leonie V.D.E. Vogelsmeier[1], Eduardo Oliveira[2], Kamila Misiejuk[3], Sonsoles López-Pernas[4], Mohammed Saqr[4]

[1] Department of Methodology and Statistics, Tilburg University, The Netherlands
[2] School of Computing and Information Systems, University of Melbourne, Australia
[3] Center of Advanced Technology for Assisted Learning and Predictive Analytics (CATALPA), FernUniversität in Hagen, Germany
[4] University of Eastern Finland, School of Computing, Joensuu, Yliopistokatu 2, 80100, Joensuu, Finland

**Declaration of interest statement**: The authors report there are no competing interests to declare.

**Data and code availability statement**: The data and analysis code that support the findings of this study are openly available in [repository name will follow] at [URL will follow].

**Disclosure of artificial intelligence-generated content (AIGC) tools**: Besides using them for data generation as explained in the Method Section, their use was limited to improving spelling, grammar, and R code for analyses.

**Author Note**: Correspondence concerning this article should be sent to either Leonie V. D. E. Vogelsmeier, Department of Methodology and Statistics, Tilburg University, Warandelaan 2, 5037 AB Tilburg, Netherlands. Email: l.v.d.e.vogelsmeier@tilburguniversity.edu.

# Abstract

Large language models (LLMs) offer the potential to simulate human-like responses and behaviors, creating new opportunities for psychological science. In the context of self-regulated learning (SRL), if LLMs can reliably simulate survey responses at scale and speed, they could be used to test intervention scenarios, refine theoretical models, augment sparse datasets, and represent hard-to-reach populations. However, the validity of LLM-generated survey responses remains uncertain, with limited research focused on SRL and existing studies beyond SRL yielding mixed results. Therefore, in this study, we examined LLM-generated responses to the 44-item Motivated Strategies for Learning Questionnaire (MSLQ; Pintrich & De Groot, 1990), a widely used instrument assessing students' learning strategies and academic motivation. Particularly, we used the LLMs GPT-4o, Claude 3.7 Sonnet, Gemini 2 Flash, LLaMA 3.1–8B, and Mistral Large. We analyzed item distributions, the psychological network of the theoretical SRL dimensions, and psychometric validity based on the latent factor structure. Our results suggest that Gemini 2 Flash was the most promising LLM, showing considerable sampling variability and producing underlying dimensions and theoretical relationships that align with prior theory and empirical findings. At the same time, we observed discrepancies and limitations, underscoring both the potential and current constraints of using LLMs for simulating psychological survey data and applying it in educational contexts.

# *Delving* into The Psychology of Machines: Exploring the Structure of Self-Regulated Learning via LLM-Generated Survey Responses

## Introduction

While artificial intelligence (AI) has been around for decades, it was not until the emergence of Generative Pre-trained Transformer (GPT) Large language models (LLMs)—particularly in 2020 with the release of GPT-3—that AI began to demonstrate human-like language abilities. More importantly, although LLMs are fundamentally complex algorithms, they have shown indications of cognitive-like abilities, such as reasoning, abstraction, and problem-solving, that in some cases resemble human performance in specific domains. Early research has shown that LLMs are "increasingly capable of simulating human-like responses and behaviors, offering opportunities to test theories and hypotheses about human behavior at great scale and speed" (Grossmann et al. 2023, p. 1108). These potentials present several opportunities. LLMs can rapidly generate large samples of human-like responses at scale, bringing diverse responses from and simulating responders that may be hard to reach at low cost (Grossmann et al. 2023). Different demographics and subgroups based on age, culture, or learning preferences can (arguably) be simulated while avoiding the risk of participant burden or ethical implications associated with data collection (Rossi et al. 2024). This becomes particularly important when studying psychological constructs such as self-regulated learning (SRL), where diverse, large-scale, and demographically varied datasets are essential for validating theoretical models, testing interventions, or applying predictive models across different learner profiles (Lee et al. 2025). For instance, LLMs can be used to explore how learners with different characteristics respond to feedback techniques or changes in learning environments, thereby informing us of the possible benefits or setbacks before applying these in real-life situations (Borisov et al., 2022; Rossi et al., 2024; López-Pernas et al., 2025).

Several challenges, including ethicality, bias, alignment with human values, and transparency, remain (Mustafa et al., 2024). First, traditional research often includes mechanisms for participants to revise or retract statements. However, LLMs shift the agency of human participants (e.g., the ability to opt out, resist, or correct misconceptions) to automated systems because LLMs are trained on large datasets composed of text generated by real people who typically did not provide explicit consent, nor the opportunity to revise or control how their information is used (Rossi et al. 2024; Stahl et al. 2022). As a result, research involving LLMs can be inherently exclusionary (Rossi et al. 2024). Furthermore, LLMs are optimized to predict the next likely word, producing outputs that prioritize linguistic plausibility over real human psychological authenticity, processes, or alignment with pedagogical goals and objectives (Sartori & Orrù, 2023). Relatedly, it is unclear whether traits like motivation or metacognition are transferred meaningfully to LLMs, or if LLM responses merely reflect surface patterns in the training data, which is far from complete, diverse, or representative. On the one hand, LLMs are notoriously acquiescent and tend to agree with most statements, which undermines the realism and variability expected in human data. Additionally, LLMs may reproduce aggregate patterns of human responses rather than the full breadth—and dare we say, also the outliers, as well as special and underrepresented populations. While LLMs may produce data that appear consistent at the aggregate level, overfitting to training data complicates interpretation—models may reproduce familiar answers rather than generate novel, persona-driven responses. On the other hand, LLMs may encode dynamic, causal structures that drive the LLM behaviors to closely resemble psychological behavior of humans that “may not be a superficial mimicry but a feature of their internal processing” (Lee et al. 2025, p. 1). Therefore, it stands to reason that researchers should further critically examine and scrutinize the internal dynamics of these models to better understand the emergence of such behaviors and their potential implications.

To assess the internal dynamics, it is crucial to rigorously evaluate the validity, reliability, and psychometric integrity of LLM-generated data to assess whether and to what extent LLMs can accurately model the complex and dynamic psychology of real learners. Establishing these characteristics—or the lack thereof—would enable us to understand LLMs' role as tools for educational research, as well as providers of guidance and support for students, given the increasing reliance on LLMs to offer advice to students, teachers, and policymakers alike. Moreover, as LLMs increasingly take on roles as tutors and surrogate instructors, understanding the psychological structure of LLMs may offer a valuable window into how they reason, deliver feedback, and shape learning experiences through their outputs.

In this study, we assess LLM-generated survey responses to an SRL questionnaire with regard to the item distributions, psychological network of theoretical constructs underlying SRL, and the psychometric validity in terms of the estimated latent factor structure. Specifically, we use the widely used Motivated Strategies for Learning Questionnaire (MSLQ; Pintrich & De Groot, 1990), which is a 44-item self-report instrument[1] designed to assess high school students’ use of learning strategies and academic motivation. MSLQ employs a 7-point Likert scale and was originally theorized to measure the five constructs *Intrinsic Value*, *Self-Efficacy*, *Test Anxiety*, *Cognitive Strategy Use*, and *Self-Regulation*. We compare the responses of five well-known frontier models: GPT-4o (OpenAI), Claude 3.7 Sonnet (Anthropic), Gemini 2 Flash (Google), LLaMA 3.1–8B (Meta), and Mistral Large (Mistral AI) (in the following, these models are abbreviated as GPT, Claude, Sonnet, Gemini, LLaMa, and Mistral). Thereby, we answer the following research questions:

RQ1: *To what extent do item response distributions generated by LLMs exhibit variability in individual item responses and theoretical construct scores?*

RQ2: *To what extent do LLM-generated datasets reproduce the psychological network structure of the MSLQ, in terms of edge connectivity, strength, directionality, and predictability of SRL processes?*

---

[1]Note that both longer and shorter versions exist; however, the 44-item version is the most commonly applied and has been validated in multiple languages, as discussed in the Properties of the Motivated Strategies for Learning Questionnaire Section. Moreover, due to memory limitations in LLMs, using longer versions would likely result in significantly more missing data, as also explained in the Data Generation Section.

RQ3: *To what extent do LLM-generated datasets replicate the latent factor structure and inter-construct relationships of the MSLQ, as evidenced by confirmatory and exploratory factor analyses?*

The study design is summarized in Figure 1. In the following, we first review prior work on LLM-generated data in survey research and explain the origins of the MSLQ. In the Method Section, we describe the 44-item version of the MSLQ, including its previous psychometric evaluations, and explain how responses were generated using LLMs. The Analysis Plan outlines all planned analytical steps, followed by the Results section, where we present the relevant findings. Finally, we conclude with a Discussion of the opportunities and threats associated with generating survey responses via LLMs, and reflect on the limitations of the current study and directions for future research.

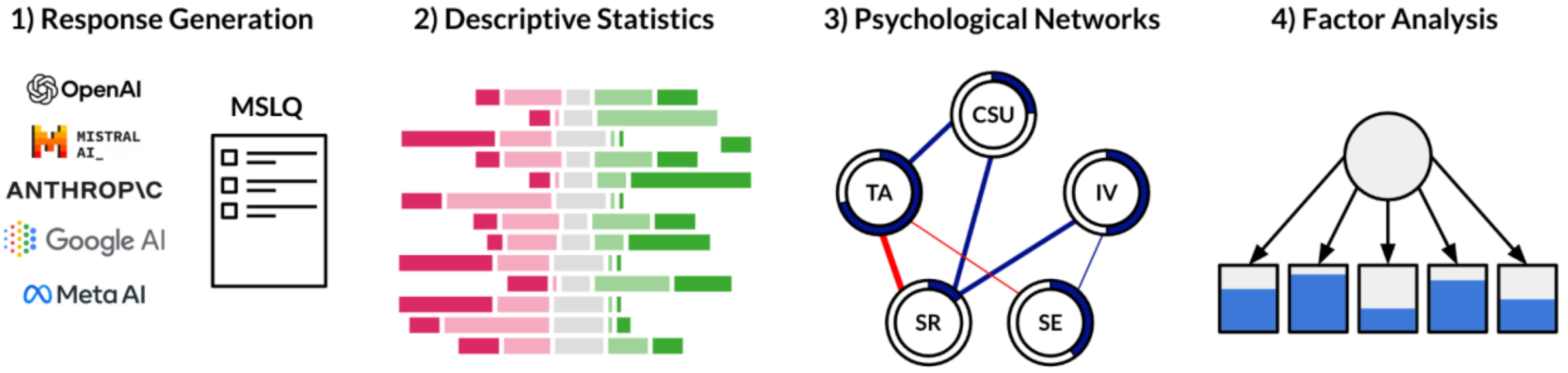

***Figure 1.*** Summary of the study design.

## Prior Work on LLM-generated Data in Survey Research

LLMs have demonstrated a wide range of capabilities, among which data generation and language in particular stand out as one of the most significant. All the more so, the data have been coherent and contextually appropriate across various domains, which included creative writing, technical documentation, and code generation. Researchers have started to explore their capabilities beyond language in generating social science data. Several studies have reported the capabilities of LLMs in generating tabular data, demonstrating their ability to produce highly realistic tabular datasets (Borisov et al., 2022). More recently, a study by López-Pernas et al. (2025) found that LLMs can generate complex network structures with properties more similar to those found in real-world data than those simulated by traditional rule-based models.

LLMs may offer several promising advantages for simulating responses to survey questionnaires by enabling large-scale deployment, reducing research costs, and generating coherent answers to open-ended questions. Compared to human participants, LLMs may also produce more consistent responses with lower variance, which can enhance the reliability of the data collected (Jansen et al., 2023). However, prior studies have highlighted the complexity and challenges associated with evaluating the quality and validity of LLM-generated survey data. For example, Bisbee et al. (2024) explored how accurately LLMs can replicate public opinion by prompting them to adopt different personas based on demographic and political interest parameters. Three LLMs (ChatGPT 4.0, ChatGPT 3.5 Turbo, Falcon-40B-Instruct) were used to generate 30 synthetic responses for each of the 7,530 respondents in the 2016–2020 American National Election Study survey. The analysis was then conducted using the average of these generated responses for each person. The findings showed that generated sampling was unreliable for statistical inference, synthetic responses showed less variation than real survey data, and regression coefficients often differed significantly from those based on human results. The study also found that the distribution of synthetic responses was sensitive to slight changes in prompt wording and varied significantly over a three-month period, even with the same prompt.

Tjuatja et al. (2024) examined whether LLMs show human-like response biases when answering survey questions. Nine LLMs (Llama2 7b, Llama2 13b, Llama2 70b, GPT 3.5 turbo, GPT 3.5 turbo instruct, Llama 7b-chat, Llama 13b-chat, Llama 70b-chat, Solar) were used to generate answers to pairs of multiple-choice questions, covering five known biases and three neutral changes. The results showed that, unlike humans, LLMs often reacted in the opposite direction of expected biases and were also sensitive to changes that should not influence responses. The authors concluded that even when a model appears to mimic the distribution of human opinions, it does not necessarily behave like a human in its sensitivity to bias-related changes.

Liu et al. (2025) evaluated six LLMs (GPT-4, GPT-3.5, Llama 3, Llama 2, Gemini, Cohere) to generate human responses together with psychometric properties to a college-level algebra assessment. The synthetic responses were compared to actual human responses and analyzed using item response theory. The results showed that the LLM-generated responses had lower variability than the human responses. Among the models, LLaMa and GPT-3.5 demonstrated above-average mathematical proficiency, with GPT-3.5 producing the most human-like response patterns in terms of psychometric characteristics.

In addition to studies exploring political opinions, response biases, and discipline-specific assessments using LLM-generated responses, several studies have been conducted to examine the feasibility of LLMs in simulating human responses to psychological questionnaires. For instance, Serapio-García et al. (2023) investigated whether LLMs can simulate personality assessments using standardized psychometric scales. The study employed the IPIP-NEO and the Big Five Inventory, prompting five LLMs (PaLM 62B, Flan-PaLM 8B, Flan-PaLM 62B, Flan-PaLM 540B, Flan-PaLMChilla 62B) to rate items independently based on the likelihood of each response option. Prompts also included persona descriptions corresponding to one of 50 demographic profiles. The results of extensive reliability, convergent and discriminant validation, and criterion validity tests indicated that Flan-PaLM 540B produced reliable and valid personality measurements. However, the authors noted that the model was primarily trained on Western data using English-language psychometric tests, potentially limiting its cross-cultural generalizability.

In a similar study, Petrov et al. (2024) examined whether GPT-3.5 and GPT-4 can simulate human-like personality traits by adopting different personas and answering standardized psychological measures. They used two types of persona descriptions: generic ones and specific ones based on demographic profiles. Each model was tested on several psychological tests, including the Big Five Inventory. GPT-4 showed somewhat human-like psychometric properties when using generic personas. However, GPT-4's responses still showed inflated correlations among traits and failed to reproduce the full Big Five factor structure. When prompted with specific demographic profiles, both models performed poorly, with weak and inconsistent psychometric results. The study concludes that current LLMs struggle to reliably simulate individual-level human behavior, especially when based on detailed demographic personas.

A recent study by de Winter et al. (2024) used ChatGPT (GPT-4) to generate 2,000 synthetic personas, each assessed using the short form of the Big Five Inventory (BFI-10). Personas were created in 100 batches of 20 using a temperature of 1 (i.e., the randomness parameter, where 0 indicates determinism and 1 promotes randomness) to encourage diversity. Each person completed the questionnaire 10 times with a temperature of 0 to ensure consistency, resulting in 20,000 total responses. Only 19 responses were excluded from the dataset due to formatting issues in the ChatGPT output, such as generating more than the expected 10 digits. Mean scores were calculated across repetitions to create the final dataset. Principal component analysis was conducted on the BFI-10 responses, along with analyses of means, standard deviations, and factor loadings. Correlation analysis showed generally plausible relationships between traits, and a clear factor structure emerged, consistent with prior research. However, some discrepancies were noted relative to findings from previous research, underscoring both the potential and the limitations of using LLMs to simulate psychological survey responses in synthetic participants. The authors suggest that employing personas could serve as a useful preliminary step for testing questionnaires and exploring research hypotheses before administering them to real human participants.

Finally, Wang et al. (2024) compared simulated responses from three LLMs (GPT-3.5, GPT-4, LLaMA 3) to two popular personality questionnaires: the Big Five Inventory-2, which includes a 5-factor model of personality, and the HEXACO-100 Personality Inventory, a six-factor model of personality. The data were generated by setting the temperature to 0 and by implementing two prompting techniques: persona prompting, in which the prompt includes five short sentences describing a person, and shape prompting, which uses five randomly selected adjectives from a specific personality domain. Approximately 300 responses were generated by each model and prompting method, and these were compared to 1,559 human responses using confirmatory factor analysis. The results indicated that LLaMa3 models more closely resembled human data in terms of factor loadings and between-factor correlations compared to other models. Newer models with a higher number of parameters showed closer alignment with human responses, suggesting that more recent models may yield improved results. The authors conclude that, despite these advances, LLMs are still not adequate substitutes for humans in psychometric research.

In this study, we extend previous research on LLM-generated survey responses by exploring their capabilities in a new domain: educational research. Specifically, we examine whether LLMs can generate complex, dynamic psychological data that reflect self-regulation processes across a diverse student population. To do this, we assess their ability to produce responses to a well-established SRL survey with a known structure, validated reliability and validity, and demonstrated utility across multiple contexts. Given the above summarized prior work on LLM-generated survey responses, we anticipate that the variability in responses may be artificially low (RQ1) (e.g., Bisbee et al., 2024; Jansen et al., 2023; Liu et al., 2025), for example, resulting in narrow item score distributions and a lack of outliers. However, also in line with the reviewed literature, we expect the generated responses to be at least somewhat aligned with the underlying theoretical constructs both in terms of the relationships between the theoretical dimensions of SLR (RQ2) and the number and nature of the underlying dimensions (RQ3) (e.g., Petrov et al., 2024; Serapio-García et al., 2023; Wang et al., 2024; Winter et al., 2024).

## Origins of the Motivated Strategies for Learning Questionnaire

SRL refers to an “active, constructive process whereby learners set goals for their learning and attempt to monitor, regulate and control their cognition, motivation, and behaviour, guided and constrained by their goals and contextual features in the environment” (Pintrich 2000, p. 453). A recent meta-analysis on SRL in higher education reported a positive correlation between overall SRL strategies and learning performance (Cheng et al., 2025). SRL can be measured in two primary ways: as an *event* or as an *aptitude*, each reflecting different conceptualizations and requiring distinct methodological approaches (Winne & Perry, 2000). Measuring SRL as an event views it as a dynamic process that unfolds over time, rather than as a fixed trait. This approach typically involves real-time methods such as think-aloud protocols, behavioral observations, or the analysis of log data as students engage with tasks. In contrast, measuring SRL as an aptitude conceptualizes it as a relatively stable mental characteristic and often relies on structured interviews or self-report surveys (Zeidner & Stoeger, 2019). In this study, we focus on self-report surveys of SRL.

Surveys are instruments used to collect data from, ideally, a representative sample of individuals from a target population to assess their attitudes, opinions, and behaviors (Taherdoost, 2016). This method is widely used across several disciplines, including psychology, the social sciences, and educational research. Survey development follows a rigorous methodology that involves multiple steps, designed to ensure the validity and reliability of the instrument. These steps help ensure that survey items accurately reflect the constructs intended by the researcher. Additionally, standardized conditions are implemented to promote consistency in responses across similar situations, thereby increasing confidence in the stability and accuracy of the collected data (Taherdoost, 2016).

One such instrument widely adopted to assess SRL is the Motivated Strategies for Learning Questionnaire (MSLQ), developed by Pintrich and De Groot (1990). The MSLQ was designed to assess classroom performance by examining the relationships between key motivational components such as intrinsic value, self-efficacy, and test anxiety, and elements of self-regulated learning, including strategy use and self-regulation. The instrument was subsequently refined and evaluated for validity and reliability in later studies (e.g., Pintrich et al., 1993). Although initially developed for college students, the MSLQ has since been validated across diverse contexts, including various disciplines, educational levels, and countries. For instance, Wang et al. (2023) tested the psychometric properties of the MSLQ in mathematics in secondary-level education for the Chinese language, Erturan İLker et al. (2014) assessed the Turkish version of MSLQ with a group of high school students, while Hilpert et al. (2013) administered MSLQ to students in introductory physical geology courses at several postsecondary institutions in the United States.

Despite its widespread adoption across various educational contexts, meta-analytic reviews have highlighted important psychometric concerns regarding the MSLQ, particularly in relation to the reliability of its scales and their predictive validity. For example, a meta-analytic review by Credé & Phillips (2011) confirmed the overall theoretical underpinning of the MSLQ but reported low correlations between its scales and academic performance, which may be attributed to poorly constructed items. Similarly, a reliability generalization meta-analysis by Holland et al. (2018) demonstrated that various characteristics of MSLQ implementation significantly influence the reliability of nearly all MSLQ subscales. Nevertheless, the MSLQ remains a prominent instrument in contemporary educational research, including in technology-enhanced learning contexts. Even in digital learning environments, where collecting digital trace data is typically the primary method of assessment, the MSLQ remains widely used to evaluate students' SRL skills. For example, a systematic review of SRL research in Massive Open Online Courses (MOOCs) by Ceron et al. (2020) found that 29 percent of the 66 included studies employed the MSLQ to collect data. Similarly, a recent meta-analysis on SRL in higher education identified the MSLQ as the most frequently used measurement tool, appearing in 16 of the 27 studies reviewed (Cheng et al., 2025). Increasingly, researchers also triangulate MSLQ survey responses with trace data to gain a more comprehensive understanding of SRL behaviors (e.g., Van Halem et al., 2020; Zhidkikh et al., 2023).

# Method

In the following section, we first describe the 44-item MSLQ and review prior psychometric evaluations to inform our expectations regarding the factor structure that the LLMs should retrieve if they have a theoretical understanding of the content. Thereafter, we detail the data generation using the five LLMs.

## Properties of the Motivated Strategies for Learning Questionnaire

According to its original conceptualization, the 44 items were designed to measure five distinct factors (Pintrich & De Groot, 1990, see Table 1): (1) *Intrinsic Value* (IV; with 9 items), (2) *Self-Efficacy* (SE; with 9 items), (3) *Test Anxiety* (TA; with 4 items), (4) *Cognitive Strategy Use* (CSU; with 13 items), (5) *Self-Regulation* (SR; with 9 items). The 44-item version of the MSLQ is a widely used questionnaire that measures student motivation and self-regulated learning, and has been evaluated over the years in various languages and cultural contexts. A summary of relevant psychometric evaluation studies is provided in Table 2. Based on prior literature evaluating the 44-item MSLQ across various languages and student populations, a 4- or 5-factor structure is most consistently supported. Studies using exploratory and confirmatory approaches (e.g., Pintrich & DeGroot, 1990; Bonanomi et al., 2018; Erturan et al., 2014) generally suggest a 5-factor model, with substantial but distinct correlations between SR and CSU (e.g., $r = .83$ and $r = .64$), supporting the retention of separate factors. In contrast, other studies (e.g., Rao & Sachs, 1999; Lee et al., 2010) propose a 4-factor model that combines SR and CSU based on weaker discriminant validity. The pattern of correlations between TA and SE also varies across studies, being more strongly negative in those supporting a 5-factor solution. Taken together, the existing evidence suggests that both a 4- and a 5-factor model are plausible, with the final structure likely depending on the specific sample and method of analysis, or, in our case, depending on the LLM used for data generation.

***Table 1.*** The Overview of the 44 Items in MSLQ

| Factor | # of items | Examples |
|---|---|---|
| Intrinsic Value (IV) | 9 | - "I think what we are learning in this Science class is interesting."<br>- "It is important for me to learn what is being taught in this English class."<br>- "I prefer class work that is challenging so I can learn new things." |
| Self-Efficacy (SE) | 9 | - "I expect to do very well in this class."<br>- "I am sure that I can do an excellent job on the problems and tasks assigned for this class."<br>- "I know that I will be able to learn the material for this class." |
| Test Anxiety (TA) | 4 | - "I am so nervous during a test that I cannot remember facts I have learned."<br>- "When I take a test I think about how poorly I am doing." |
| Cognitive Strategy Use (CSU) | 13 | - "When I read material for science class, I say the words over and over to myself to help me remember."<br>- "When I study for this English class, I put important ideas into my own words."<br>- "I outline the chapters in my book to help me study." |
| Self-Regulation (SR) | 9 | - "I ask myself questions to make sure I know the material I have been studying."<br>- "I find that when the teacher is talking I think of other things and don't really listen to what is being said."<br>- "Even when study materials are dull and uninteresting, I keep working until I finish." |

*Note*. The complete list of items can be found in Pintrich & DeGroot (1990).

***Table 2.*** Literature evaluating the 44-item MSLQ

| Nr. | Short Reference | Language of Questionnaire | Number of Factors Suggested | Psychometric Assessment |
|---|---|---|---|---|
| 1 | Pintrich & DeGroot, 1990 | American English | 5 (correlation SR/CSU = .83) | EFA |
| 2 | Bonanomi et al., 2018 | Italian | 5 (correlation SR/CSU = .64) | Multidimensional Rasch analysis |
| 3 | Rao and Sachs, 1999 | Chinese | 4 (SR and CSU combined) | CFA |
| 4 | Lee et al., 2010 | Chinese | 5 or 4 (SR and CSU combined) | Multidimensional Rasch analysis |
| 5 | Liu et al., 2012 | Singapore | 8 (SR and CSU distributed across multiple scales) | PCA |
| 6 | Erturan et al., 2014 | Turkish | 5 (correlation SR/CSU = unclear) | CFA |

*Notes*. The population was always high school students. Only #2 used a 1-5 Likert scale; all others used the by #1 proposed 1-7 scale. The main differences in findings lie in the correlations between SR and CS, which are substantial in #1 and #2 but not strong enough to justify combining the factors, as was done in #3 and #4. Additionally, the correlation between TA and SE is notably negative in #1 and #2, whereas it is weak or absent in #3 and #4. EFA = Exploratory Factor Analysis. CFA = Confirmatory Factor Analysis. PCA = Principal Component Analysis.

## Data Generation

To generate large-scale, demographically diverse responses to the MSLQ, we developed a structured, prompt-based pipeline using five large language models (LLMs): GPT-4o (OpenAI), Claude 3.7 Sonnet (Anthropic), Gemini 2 Flash (Google), LLaMA 3.1–8B (Meta), and Mistral Large (Mistral AI). These models represent a diverse sample of popular LLMs, ranging from proprietary, high-performing options like GPT-4o and Claude 3.5 Sonnet to open-source models such as LLaMA 3.1–8B, which can even be run on local machines. They vary significantly in size, architecture, and accessibility. Some models are optimized for speed and low latency, while others are optimized for better reasoning or flexibility in deployment. Our rationale for including this selection in our research is to compare how these popular and different types of models generate synthetic data in the context of our study.

For each model, we simulated responses from 1,000 unique students, with each student completing all 44 items from the MSLQ on a 7-point Likert scale. In addition to item-level ratings, each simulated student was assigned a coherent demographic profile by an LLM, including full name, age, educational background, country, ethnicity, disability status, and study area. As mentioned in the introduction, we opted to use the 44-item version of the MSLQ instead of the full 81-item questionnaire. While the full version offers broader psychometric coverage, it would have significantly increased the length of both the prompt and the model's output, raising the likelihood of token truncation, prompt overflow, and silent failures across different LLM APIs. These issues can introduce artificial missingness, reducing the reliability and interpretability of simulated psychometric data. Our own experiments support this concern: even with the 44-item version, model responses often began to deviate from the intended direction toward the end of the questionnaire. Thus, relying on the 81 version would have made it very difficult to obtain the sample from all models, especially LLaMa, which has issues with missing data. The 44-item version, by contrast, is widely validated and frequently cited in educational research, and also focuses on important subscales. For the purposes of this study, centered on construct-level coherence and psychometric structure across synthetic datasets, it offered a strong balance between theoretical depth, practical feasibility, and representativeness of all models. All technical details about the data generation are provided in Appendix A.

# Analysis Plan

In this Section, we outline the analysis steps in detail, corresponding to steps 2 through 4 in Figure 1. We begin with data cleaning, followed by descriptive statistics. Afterwards, we proceed to the construction of psychological networks and the psychometric validation.[2]

## Data Cleaning

After collecting the data, we first clean the dataset by addressing issues that may arise, such as repeated participant names and incoherent responses (e.g., text instead of numeric values).

[2] Note that the data is expected to have incomplete cases as explained in the Data Generation Section. For the network and psychometric analyses, we use complete-case analyses. In Appendix C, we explain why this is warranted for the specific case of LLM-generated data.

Since specific problems only become apparent upon inspection, we cannot predefine all remedies. However, we document the cleaning steps transparently in the Results Section.

## Descriptive Statistics

### Central Tendency and Variability

To examine the output of different LLMs in simulating responses, we compute descriptive statistics (mean and standard deviation) for the MSLQ responses provided for each LLM at two levels of granularity: at the item level and at the theoretical construct level. Prior to computing construct-level scores, four items with reverse-scored wording (26, 27, 37, and 38) are recoded to ensure alignment with the directionality of the construct (e.g., a 1 became a 7, a 2 became a 6, etc.). The resulting descriptive statistics are used to compare how different LLMs represent motivational and learning strategies. We also plot the response distribution using the R package *likert* (Bryer & Speerschneider, 2016) to visually identify differences in patterns of responses.

### Country Proportions

Given that LLMs were primarily trained on English texts and Western-centric data (e.g., Liu et al., 2024; Cao et al., 2023), we examine the country proportions to determine whether the sample is disproportionately represented by participants from the United States. Such an imbalance could potentially explain the emergence of four versus five factors in the psychometric validation, as prior empirical studies have typically identified four factors in Asian samples and five in American ones (see Properties of the Motivated Strategies for Learning Questionnaire Section).

### Outliers

Moreover, given the prior work on generally low variability in LLM-generated responses (e.g., Jansen et al., 2023; Liu et al., 2025), we assess whether the LLM-generated datasets contain outliers. We do not examine individual item scores because, with a 7-point Likert scale, only illegal values are true outliers at the item level. Instead, we use sum scores per theoretical dimension, which allows us to detect construct-level outliers and assess meaningful construct variability in the data. Specifically, we use the median absolute deviation (MAD) method with a threshold of 3. Note that we consider only complete data, as sum scores based on fewer than 44 responses would be automatically deviant. This outlier check will be complemented by a factor analysis model-based outlier check at the observation level, as will be explained in the Model-Based Outlier Section. As mentioned before, it is important to note that we do not have a dataset containing responses from real human participants, and therefore, no direct comparison is possible. However, if the variability in the simulated data is artificially low, we would not expect to detect any outliers.

## Psychological Networks

To further investigate the structure of the responses, we represent the relationship between the different theoretical constructs in the MSLQ questionnaire as psychological networks, where the nodes correspond to constructs (i.e., IV, SE, TA, CSU, SR), and the edges represent conditional associations between them after controlling for all other constructs in the network.

The networks offer valuable insights into the underlying structure of the data, the interactions between the SRL processes, magnitude, direction, and explainability. While there is no ground-truth model, it is natural that all SRL processes will be associated with some negative correlations with TA (Bonanomi et al., 2018; Birenbaum & Dochy, 2012; Rao & Sachs, 1999; Wang, 2012). The associations between constructs are estimated as partial correlations using a Gaussian Graphical Model (GGM) with regularization, implemented via the EBICglasso method in the *bootnet* R package (Epskamp et al. 2018). Two constructs are partially correlated when they are conditionally dependent on each other after controlling for all other constructs in the network (similar to regression). Furthermore, we calculate the predictability of each construct ($R^2$), which represents the proportion of variance in a given construct that can be explained by all other constructs directly connected to it, together with the prediction error (root mean squared error; RMSE). This is done via the *mgm* R package (Haslbeck and Waldorp 2020). The networks are represented using *qgraph* (Epskamp et al., 2012), where the nodes represent the constructs, the donut chart surrounding the nodes indicates the $R^2$, and the edges between the nodes represent the partial correlations.

## Psychometric Validation

To psychometrically validate the generated data, we explore the underlying factor structure of five LLMs using exploratory factor analysis (EFA), following the methodology and guidelines outlined in the tutorial on factor analyses for learning analytics by Vogelsmeier et al. (2024). For a (non-technical) recap of EFA, we refer the reader to this book chapter. The results are reviewed by two domain experts: one in psychology and psychometrics, and another in artificial intelligence and machine learning. Before conducting the EFAs, we evaluate the suitability of the data for factor analysis in terms of item correlations, as detailed below. Additionally, we use confirmatory factor analysis (CFA) to assess how well both the originally proposed 5-factor model and the 4-factor model, previously identified in Asian contexts, fit the data.

### Item Correlations

We assess whether correlations between items are sufficiently strong. First, we examine heatmaps of item correlations using the R package *pheatmap* (Kolde, 2019). These visualizations represent individual correlation values through color intensity, making it easier to detect patterns and potential clusters than by inspecting numerical values alone. This provides an initial intuition about the relationships among items and the possible common factors underlying them. Furthermore, we conduct Bartlett's test of sphericity, which tests whether the item correlation matrix is an identity matrix. A rejection of the null hypothesis indicates that there are meaningful correlations between items, justifying the use of factor analysis. Additionally, we calculate the Kaiser-Meyer-Olkin (KMO) statistic, which gauges the proportion of total variance among variables that can be attributed to common variance. A KMO value of at least 0.8 is considered acceptable (though this is a rough guideline, and generally, higher values are preferred). The R package *psych* (Revelle, 2024) is used for both analyses.

## Factor Analyses

### *Confirmatory Factor Analyses*

After completing these initial steps, we apply CFA to each of the datasets using the R package *lavaan* (Rosseel, 2009), testing models with both five and four factors, as existing evidence suggests that both models are plausible (see literature research in the Properties of the Motivated Strategies for Learning Questionnaire Section). To assess model fit, we follow the recommendations of Vogelsmeier et al. (2024), evaluating whether the Chi-Square test statistic is non-significant at an alpha level of 0.05, the Comparative Fit Index (CFI) exceeds 0.90, the Root Mean Square Error of Approximation (RMSEA) is below 0.05, and the Standardized Root Mean Square Residual (SRMR) is under 0.08. We consider the model to demonstrate adequate global fit if at least two of these criteria are met.

### *Exploratory Factor Analyses*

Next, we proceed with the EFAs on the data deemed suitable based on the preliminary checks. First, using the *psych* package (Revelle, 2024), we conduct a parallel analysis to gain an initial understanding of the suggested number of factors. Next, using the R package *lavaan* (Rosseel, 2009), we perform EFAs with up to 10 factors, which is double the number of factors suggested by theory. We then examine the Bayesian Information Criterion (BIC) to identify the top three models. However, in practice, it is common for the BIC to continue decreasing as the model complexity increases (e.g., McNeish & Harring, 2017). If this happens, we place greater emphasis on the results from the parallel analysis and, ultimately, on the interpretability of the factors. In examining the factor loadings, we consider loadings substantial if their absolute values exceed 0.30. Furthermore, we allow the factors to correlate by applying an oblique rotation, which is more appropriate given the theoretical and empirical basis of the MSLQ (e.g., Rao & Sachs, 1999). If the true correlations among factors are negligible, this will be reflected in low interfactor correlations in the results. The R code used for the analyses is shared as an online supplement.

### *Model-Based Outliers*

As a final step, we explore model-based outliers in the EFAs using the forward search algorithm (Mavridis & Moustaki, 2008) implemented in the *faoutlier* package (Chalmers & Flora, 2015). For a specified number of factors, the algorithm begins by generating multiple random subsets of the data and selects the most homogeneous base subset using a minimum likelihood criterion. It then iteratively adds observations based on the Goodness-of-Fit and Mahalanobis distance, allowing for sequential assessment of case influence and the detection of outliers. Specifically, one can examine the generalized Cook's Distance (gCD) values for all observations added after the initial base subset and assess whether any stand out as unusually large and hence influential. We apply this approach to further assess whether LLM-generated data exhibit sample variability, including outliers, or if they merely reflect values close to population means, which one might expect in real-world data.

# Results

## Data Cleaning

As explained in the Data Generation Section, each of the five LLMs was queried to produce 1,000 responses. The data were requested from all LLMs through APIs and web interfaces when possible. The initial dataset for all five LLMs consisted of 263,050 rows, as responses were stored in long format. The following data problems were identified: repeated names, missing IDs, absent question numbers, and incoherent responses (text instead of numbers). We first excluded the missing question numbers, resulting in a total dataset of 260,071 records. To disambiguate repeated names, a combination of name, ID, LLM, and other demographic information was used to distinguish unique entries. However, some entries were identical regarding all identifying information and demographics; therefore, we used the sequence of questions to distinguish between successive entries. The data was then changed from long to wide format (i.e., having one column per questionnaire indicator). The final sample sizes (and hence rows) per LLM were as follows: Gemini had 882 responses, GPT had 984, LLaMa had 989, Mistral had 607, and Claude had 893.

As a final step, we checked for illegal values. LLaMa exhibited a small number of illegal values above the maximum (scores of 8 and 9) for all items. These invalid responses were excluded, resulting in a remaining sample of 965 valid observations. In the case of Mistral, one extreme illegal value (71) was recorded for the first item and was likewise excluded from further analyses, resulting in 606 valid observations.

## Descriptive Statistics

### Central Tendency and Variability

In the following, we report means and SDs aggregated across items belonging to the same theoretical dimension, omitting item-level statistics to avoid redundancy and visual overload, as patterns at the item level closely mirror those at the dimension level. However, item-level means and SDs are provided in Appendix B (Table B1), and item distributions are shown in Appendix B (Figure B1). As can be seen in Table 3, and considering the item score range of 1–7, means across all dimensions and LLMs indicate a modest tendency toward agreement with the item content. SDs vary considerably between LLMs: Gemini and LLaMa display SDs above 1.5, suggesting notable response variability. In contrast, GPT and, to a lesser extent, Mistral and Claude show consistently lower SDs, indicating more stable response patterns across items within each dimension.

***Table 3***. Means and SDs Across Items of the Five Theoretical Dimensions of the MSLQ by LLM

| | Gemini | | GPT | | LLaMa | | Mistral | | Claude | |
|---|---|---|---|---|---|---|---|---|---|---|
| **Dimension** | **M** | **SD** | **M** | **SD** | **M** | **SD** | **M** | **SD** | **M** | **SD** |
| CSU | 4.61 | 1.59 | 5.68 | 0.56 | 4.94 | 1.29 | 4.96 | 0.76 | 4.97 | 0.68 |
| IV | 5.55 | 1.64 | 6.07 | 0.55 | 5.10 | 1.50 | 5.55 | 0.85 | 5.92 | 0.82 |
| SE | 4.82 | 1.59 | 5.89 | 0.60 | 5.04 | 1.47 | 4.95 | 0.81 | 4.93 | 0.93 |
| SR | 4.89 | 1.78 | 5.52 | 0.52 | 4.56 | 0.84 | 4.97 | 0.74 | 5.05 | 1.01 |
| TA | 3.50 | 1.85 | 3.73 | 0.66 | 4.31 | 1.54 | 3.71 | 0.81 | 4.50 | 1.11 |

### Country Proportions

Table 4 presents the percentages of continents overall and across the five LLMs. Focusing on North America and Asia (because these are the areas in which the MSQ has mainly been psychometrically assessed), the percentage for North America is generally higher (54.50%) than that for Asia (23.20%). This disparity suggests that the theoretical 5-factor structure might be more plausible if the machine-generated data leans more heavily on USA data, potentially reflecting a more substantial influence from USA-centric training sources. There are no major differences across LLMs, as the percentages are close to the overall values. Only Mistral shows a noticeably higher percentage for North America (73.20%) compared to the mean, while GPT shows a noticeably lower percentage (39.90%).

***Table 4.*** Frequency and Percentages of the Continents to Which the Countries Generated by the LLMs Belong, Overall and by LLM

| Continent | Overall | Claude | Gemini | GPT | LLaMa | Mistral |
|---|---|---|---|---|---|---|
| North America | 1920 (54.50%) | 420 (55.20%) | 424 (57.30%) | 338 (39.90%) | 392 (55.80%) | 346 (73.20%) |
| Asia | 819 (23.20%) | 136 (17.90%) | 176 (23.80%) | 241 (28.50%) | 157 (22.30%) | 109 (23.00%) |
| Europe | 345 ( 9.80%) | 84 (11.00%) | 19 (2.60%) | 139 (16.40%) | 87 (12.40%) | 16 (3.40%) |
| Africa | 235 ( 6.70%) | 52 (6.80%) | 114 (15.40%) | 53 (6.30%) | 16 (2.30%) | 0 (0.00%) |
| Oceania | 126 ( 3.60%) | 46 (6.00%) | 3 (0.40%) | 41 (4.80%) | 36 (5.10%) | 0 (0.00%) |
| South America | 41 ( 1.20%) | 2 (0.30%) | 3 (0.40%) | 20 (2.40%) | 14 (2.00%) | 2 (0.40%) |
| Europe/Asia | 38 ( 1.10%) | 21 (2.80%) | 1 (0.10%) | 15 (1.80%) | 1 (0.10%) | 0 (0.00%) |

### Outliers

Table 5 displays the number of outliers per sum score for each dataset, indicating that the number of MAD-based outliers varies substantially across LMMs and dimensions. Gemini shows a considerable number of outliers in the IV dimension. In contrast, LLaMa has no outliers, Claude has almost no outliers besides the IV dimension, and GPT and Mistral have very few outliers overall.

***Table 5.*** Number of MAD-Based Outliers per Theoretical Construct by LLM

| Dimension | Gemini | GPT | LLaMa | Mistral | Claude |
|---|---|---|---|---|---|
| CSU | 0 | 2 | 0 | 6 | 1 |
| IV | 98 | 9 | 0 | 26 | 39 |
| SE | 10 | 6 | 0 | 30 | 0 |
| SR | 1 | 8 | 0 | 8 | 0 |
| TA | 17 | 12 | 0 | 23 | 0 |

*Note*. Only complete cases were considered.

## Psychological Networks

The networks are depicted in Figure 2. In line with theory, TA was negatively associated with SR in the five models, with varying magnitudes. It was also negatively correlated with SE in three models (Gemini, Claude, and Mistral). Interestingly, TA was strongly and positively correlated with CSU in Mistral, Claude, and Llama, and absent in Gemini. In all the models, CSU was associated with SR, as well as SE and IV. In Gemini, there was an absent association between IV and SE, while the same edge was negative in Claude.

We also used the predictability to examine the connections of each node and assess how the connections thereof explain it. In all the models and across all values, $R^2$ was very high for all nodes except for GPT. The average predictability was highest in Gemini (0.92), followed by Mistral (0.85), LLaMa (0.84), Claude (0.75), and lowest in GPT (0.68), given its low value for TA. Similarly, the RMSE values ranged from 0.27 in Gemini to 0.54 in GPT, which are relatively low values. Given that the response variable was measured on a 1–7 scale, the observed RMSE values represent relatively low prediction errors. When expressed as a proportion of the full response range (6 units), the RMSE values correspond to approximately 4.5% to 9% of the total scale, suggesting that the networks demonstrated good overall predictive accuracy. These values are, in fact, far higher than those found in real-life data, which may indicate overfitting. Real-life data is often noisy, inherently varying with outliers and sometimes careless responses. Such high $R^2$ values across all nodes may obscure the model's ability to differentiate between central and peripheral processes. If every node is nearly perfectly predicted by its neighbors, it becomes difficult to infer which processes are driving the system versus being driven by it.

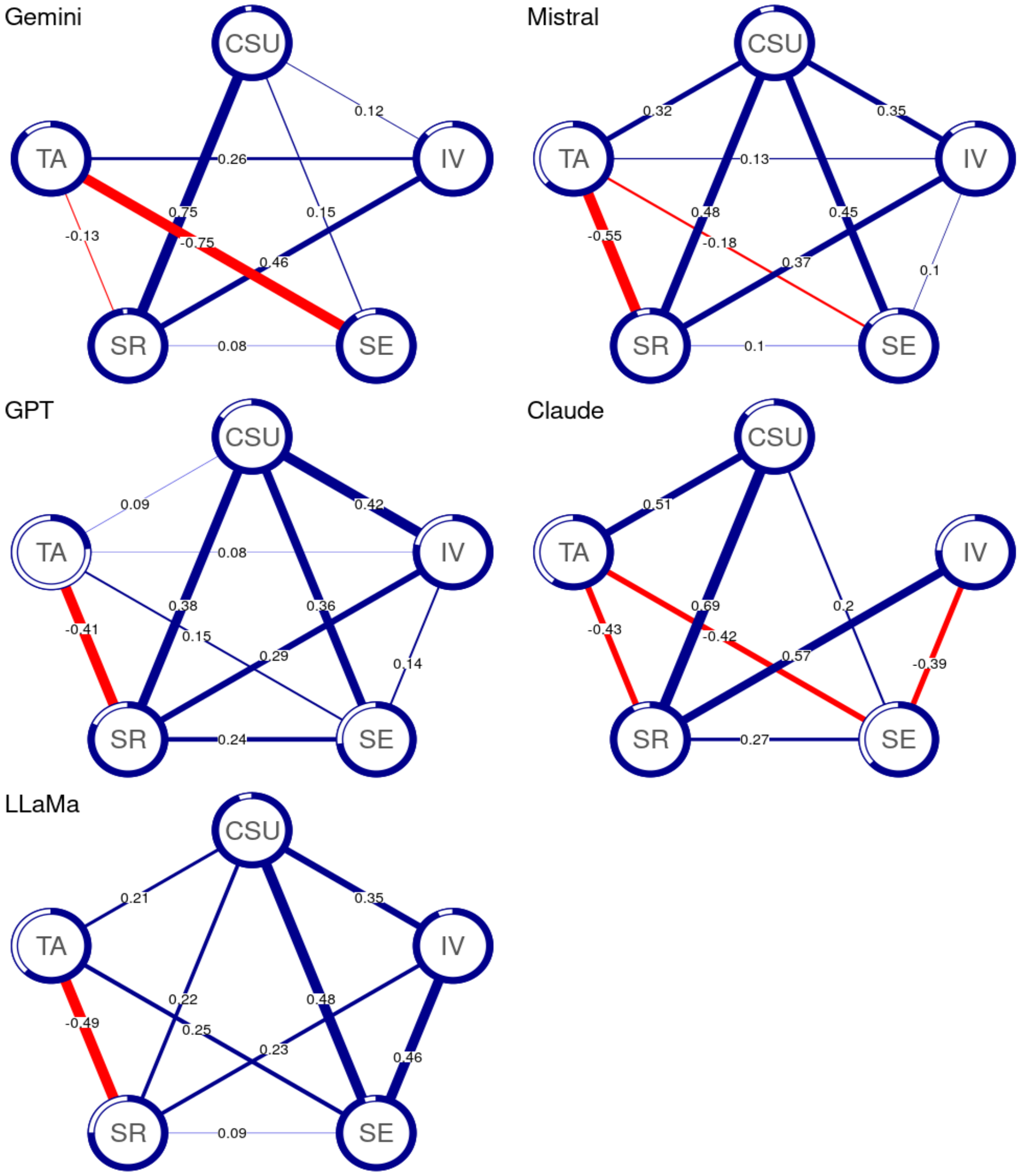

***Figure 2.*** Psychological networks of the five theoretical dimensions of the MSLQ by LLM.

***Table 6.*** Explained Variance and Prediction Errors for the Psychological Networks of the Five Theoretical Dimensions of the MSLQ by LLM

| | Gemini | | GPT | | LLaMa | | Mistral | | Claude | |
|---|---|---|---|---|---|---|---|---|---|---|
| **Dimension** | **RMSE** | **$R^2$** | **RMSE** | **$R^2$** | **RMSE** | **$R^2$** | **RMSE** | **$R^2$** | **RMSE** | **$R^2$** |
| CSU | 0.178 | 0.968 | 0.397 | 0.842 | 0.242 | 0.941 | 0.255 | 0.935 | 0.376 | 0.859 |
| IV | 0.358 | 0.871 | 0.467 | 0.782 | 0.256 | 0.934 | 0.341 | 0.884 | 0.493 | 0.756 |
| SE | 0.306 | 0.906 | 0.522 | 0.728 | 0.229 | 0.947 | 0.376 | 0.858 | 0.608 | 0.630 |
| SR | 0.154 | 0.976 | 0.426 | 0.819 | 0.501 | 0.749 | 0.247 | 0.939 | 0.280 | 0.921 |
| TA | 0.345 | 0.881 | 0.876 | 0.232 | 0.618 | 0.617 | 0.611 | 0.626 | 0.639 | 0.591 |
| Mean | 0.268 | 0.920 | 0.538 | 0.681 | 0.369 | 0.838 | 0.366 | 0.848 | 0.479 | 0.751 |

## Psychometric Validation

In the following, we present our analysis of the data's suitability for factor analysis, including correlation structures and the results of Bartlett's test and the KMO diagnostic for all five LLMs. After confirming suitability, we present the results of the EFA.

### Item Correlations

The correlation heatmaps (Figure 3) for the five LLMs reveal distinct patterns that suggest varying underlying latent structures. In these heatmaps, color intensity represents the strength of correlation between items, with blue indicating high positive correlations and red indicating low or negative correlations. Gemini and Mistral reveal clear, block-like clusters of high correlations, suggesting well-defined latent factors and supporting the use of EFA. In contrast, GPT and LLaMa exhibit smoother gradient patterns, pointing to a potential dominant general factor or a continuum rather than distinct constructs. The heatmap for Claude shows a more mixed and less structured pattern, implying overlapping or less clearly separable latent dimensions.

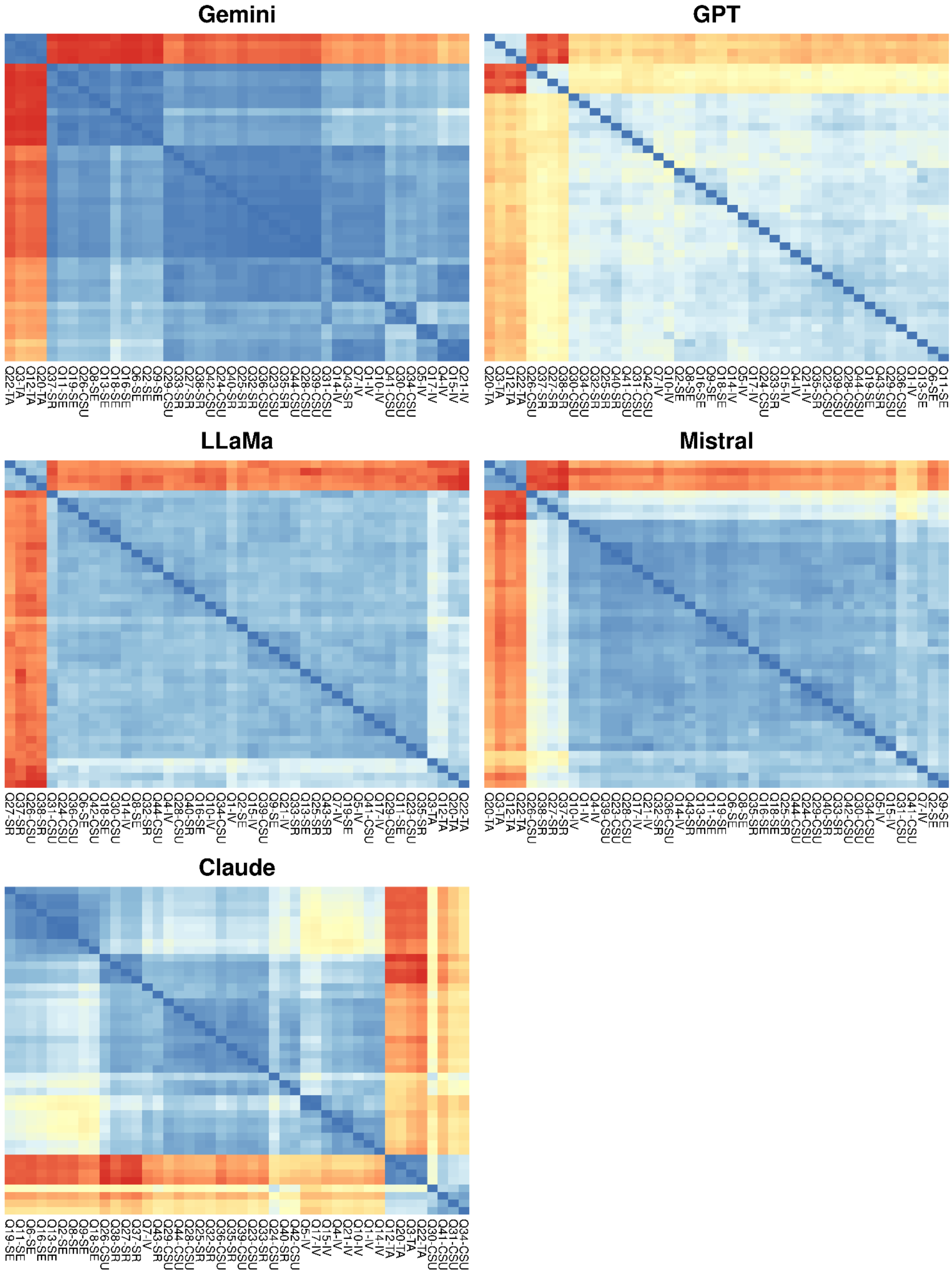

***Figure 3.*** Pairwise correlation heatmaps of response patterns across five LLMs, displaying the correlation between item responses, with darker blue indicating higher similarity and red/yellow indicating divergence. Rows and columns are arranged to group similar items together, revealing structural patterns. Items 26, 27, 37, and 38 have reversed coding.

## Factor Analyses

### *Confirmatory Factor Analyses*

The results are presented in Table 7. Neither the original 5-factor structure nor the subsequently proposed 4-factor structure (which combines CSU and SR) provided a good fit for the data from any of the five LLMs. For Gemini, GPT, and LLaMa, only the SRMR met the model fit criteria in both the 4- and 5-factor models. In contrast, for Mistral and Claude, none of the four fit indices supported an adequate model fit for either factor structure. The EFAs reported next help to explain the potential reasons behind this lack of fit.

***Table 7.*** Global fit for the Five LLMs when Applying CFA with Both 5 and 4 Factors

| | Chi-Square | df | *p*-value | CFI | RMSEA | SRMR |
|---|---|---|---|---|---|---|
| | | | ***5 factors*** | | | |
| Gemini | 13157.89 | 892 | <0.01 | 0.81 | 0.14 | **0.06** |
| GPT | 4788.59 | 892 | <0.01 | 0.83 | 0.07 | **0.07** |
| LLaMa | 5978.06 | 892 | <0.01 | 0.71 | 0.10 | **0.07** |
| Mistral | 7430.57 | 892 | <0.01 | 0.78 | 0.12 | 0.08 |
| Claude | 13365.67 | 892 | <0.01 | 0.75 | 0.14 | 0.13 |
| | | | ***4 factors*** | | | |
| Gemini | 13207.87 | 896 | <0.01 | 0.81 | 0.14 | **0.06** |
| GPT | 4796.07 | 896 | <0.01 | 0.83 | 0.07 | **0.07** |
| LLaMa | 6002.22 | 896 | <0.01 | 0.71 | 0.10 | **0.07** |
| Mistral | 7495.07 | 896 | <0.01 | 0.78 | 0.12 | 0.09 |
| Claude | 13428.71 | 896 | <0.01 | 0.75 | 0.14 | 0.13 |

*Note.* The values that align with the criteria for a good model fit are in boldface.

### *Exploratory Factor Analyses*

Parallel analysis chose 4 factors for Gemini, 6 factors for GPT, 7 for LLaMa, and 5 factors for Mistral and Claude. For all LLMs but GPT, the BIC values continued to increase with each additional factor, up to the maximum specified number of 10 factors. As explained in the Analysis Plan Section, we therefore continued with the parallel analysis recommendations. For GPT, the best three models were 7, 6, and then 8. Since 6 was suggested by parallel analysis and it is closer to the theoretical 5 factors, we choose 6.

Figure 4 visualizes the factor loadings for the selected number of factors using a color scheme (for the loading values, see Appendix D). The items are grouped by five psychological constructs as indicated by the labels on the right. Each of the five panels corresponds to one LLM and shows factor loadings across up to six extracted factors (f1–f7, shown on the x-axis). Using the theoretical groupings as panel dividers allows a visual comparison between the theoretical structure and the empirically estimated one. This facilitates assessing whether items within a given construct tend to load together on the same factor within each LLM. The bubbles represent the loading of each item onto each factor, where color indicates the direction and magnitude of the loading (blue = strong positive, red = strong negative, yellow = near-zero) and size reflects the absolute value of the loading (larger = stronger). Importantly, the sign of factor loadings is arbitrary within each factor; that is, re-estimating the model could result in all signs flipping. As such, both large blue and large red bubbles indicate strong factor loadings, and both reflect meaningful associations between the item and the latent factor. Additionally, items Q26 (belonging to CSU), Q27, Q37, and Q38 (all three belonging to SR) are reverse-coded, which should lead to opposite-sign loadings if correctly recognized by the LLMs.

The models Claude and Gemini exhibit clearer and more distinct factor structures, with items within constructs tending to load strongly onto specific factors. Furthermore, the reverse-coded items display larger red bubbles (while the regularly coded items display larger blue bubbles), which is consistent with the correct recognition of their reversed framing and supports the validity of the factor structure. In contrast, GPT, LLaMa, and Mistral show more diffuse patterns, suggesting a broader or less differentiated interpretation of the constructs. These three LLMs also fail to group the reverse-coded items with their intended construct, suggesting a weaker grasp of the item's directionality or underlying meaning.

In our comparison with the theoretical structures, we focus on the two most promising LLMs, Claude and Gemini. For Claude, the estimated 5-factor structure generally combines CSU, IV, and SR into a single factor, possibly because all these factors are approaches to understanding and remembering material. The items thus may reflect a broader underlying factor, such as "learning engagement". However, four CSU items deviate from this pattern: Q41 ("*When I read materials for this class, I say the words over and over to myself to help me remember.*"), Q34 ("*When I study for a test I practice saying the important facts over and over to myself.*"), Q31 ("*When studying, I copy my notes over to help me remember material.*"), and Q30 ("*When I study for a test I try to remember as many facts as I can.*") with loadings on the TA factor. Since the items pertain to a test situation (or to remembering content likely for that purpose), it is reasonable that they are associated with TA. Additionally, two IV items clearly deviate from the overall pattern: Q5 ("*I like what I am learning in this class.*") and Q17 ("*I think that what we are learning in this class is interesting.*"). These items exhibit cross-loadings on a factor where only they load strongly, which may represent a subtype of intrinsic value related to interest. Another three IV items show somewhat weaker deviations: Q4 ("*It is important for me to learn what is being taught in this class.*"), Q21 ("*Understanding this subject is important to me.*"), and Q15 ("*I think that what I am learning in this class is useful for me to know.*") have small cross-loadings on a separate factor that could be labeled as a subtype of intrinsic value, emphasizing importance and usefulness.

For Gemini, only four factors were extracted, but the results closely resemble those for Claude. Once again, CSU, IV, and SR are combined into a single factor. Of the four CSU items that deviated in Claude, three (Q41, Q34, and Q30; see previous paragraph for content) also deviate here, though they now exhibit only cross-loading on an additional factor, still primarily loading with the other CSU items. A notable difference from the Claude results is that the reverse-coded item Q26 ("*It is hard for me to decide what the main ideas are in what I read.*") loads with the TA items on the TA factor. Since understanding main ideas is crucial for exams, it is reasonable to associate this item with test anxiety.

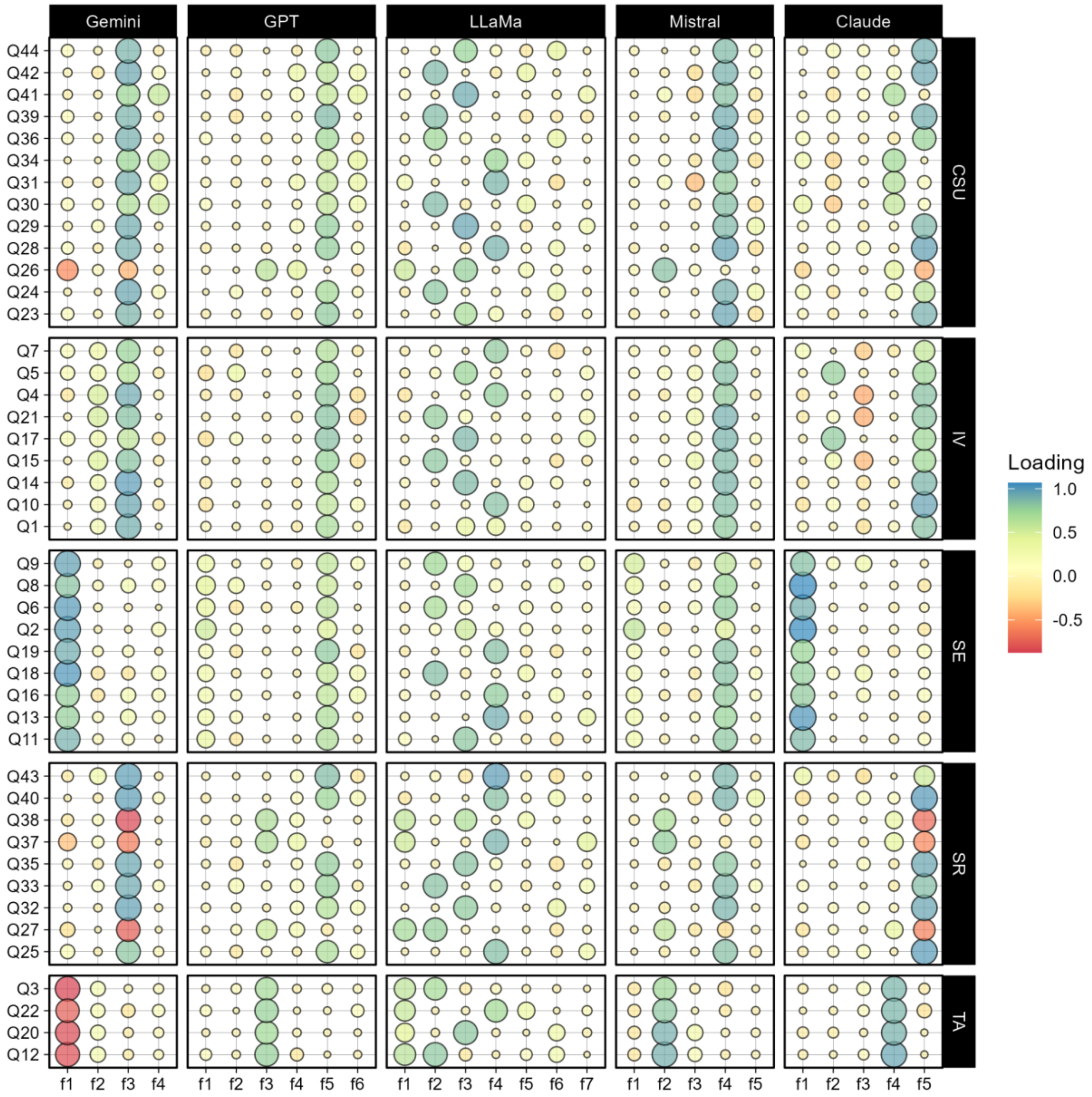

***Figure 4.*** Factor loadings from the EFA, with columns representing extracted factors (f1 – f7 on the x-axis) and rows grouped by theoretical constructs (CSU, IV, SE, SR, TA). Bubble size indicates loading strength; color shows direction (blue = positive, red = negative). Items 26, 27, 37, and 38 have their original coding (not reversed) and thus should show loadings in the opposite direction to the regularly coded items within the same construct.

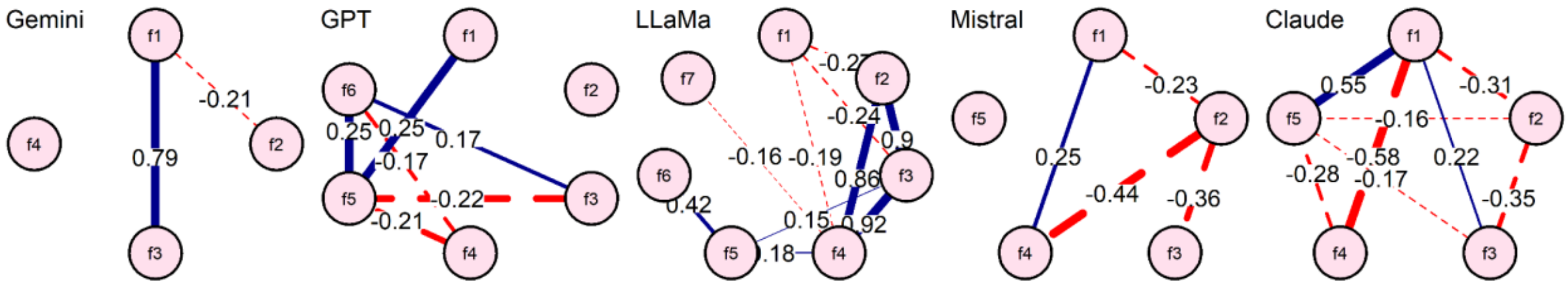

***Figure 5.*** Correlations between the estimated factors. To increase readability, we depict only loadings larger than 0.15. Which factors (f1–f7) represent which (combinations of) theoretical dimensions can be derived from the presence and absence of strong loadings in Figure 4.

Figure 5 shows the correlations between the estimated factors. Again, we focus on the two most promising LLMs, Claude and Gemini. Examining the correlations between the most meaningful estimated factors discussed above, we observe that the factors representing TA are negatively related to the other (combinations of) estimated factors, which is in line with previous empirical findings that, if a relation exists, it is negative. For Gemini, this is visible from the correlation between f1 (TA, SE) and f3 (CSU, IV, SR); although the arrow is blue and hence positive (r = .079), the loadings are all negative for TA and, thus, we can flip the signs of both the loadings and the correlations involving TA. Moreover, since TA and SE load on the same factor but with opposite signs, we can also conclude that these two constructs are highly negatively correlated. For Claude, we observe negative correlations between f4 (TA) and f5 (CSU, IV, SR), as well as between f4 (TA) and f1 (SE) (r = -0.28 and r = -0.58, respectively). Furthermore, we observe that the factors representing SE were positively correlated with the factor representing the other three dimensions, which aligns with previous empirical findings. For Gemini, f1 (TA, SE) and f3 (CSU, IV, SR) are positively related (r = .079), and in Claude, f1 (SE) and f5 (CSU, IC, SR) are positively related (r = 0.22).

### *Model-Based Outliers*

Figure 4 shows the gCDs at each step of the forward search algorithm. The results reveal notable spikes in gCDs across all models, indicating the presence of influential observations, which can be considered outliers in the context of factor analysis. Among the models, Claude exhibited a single extreme outlier that dominated the scale; once this outlier was removed, additional outliers became visible. LLaMa and Mistral showed frequent moderate outliers, whereas GPT and Gemini displayed fewer, though still present, outliers.

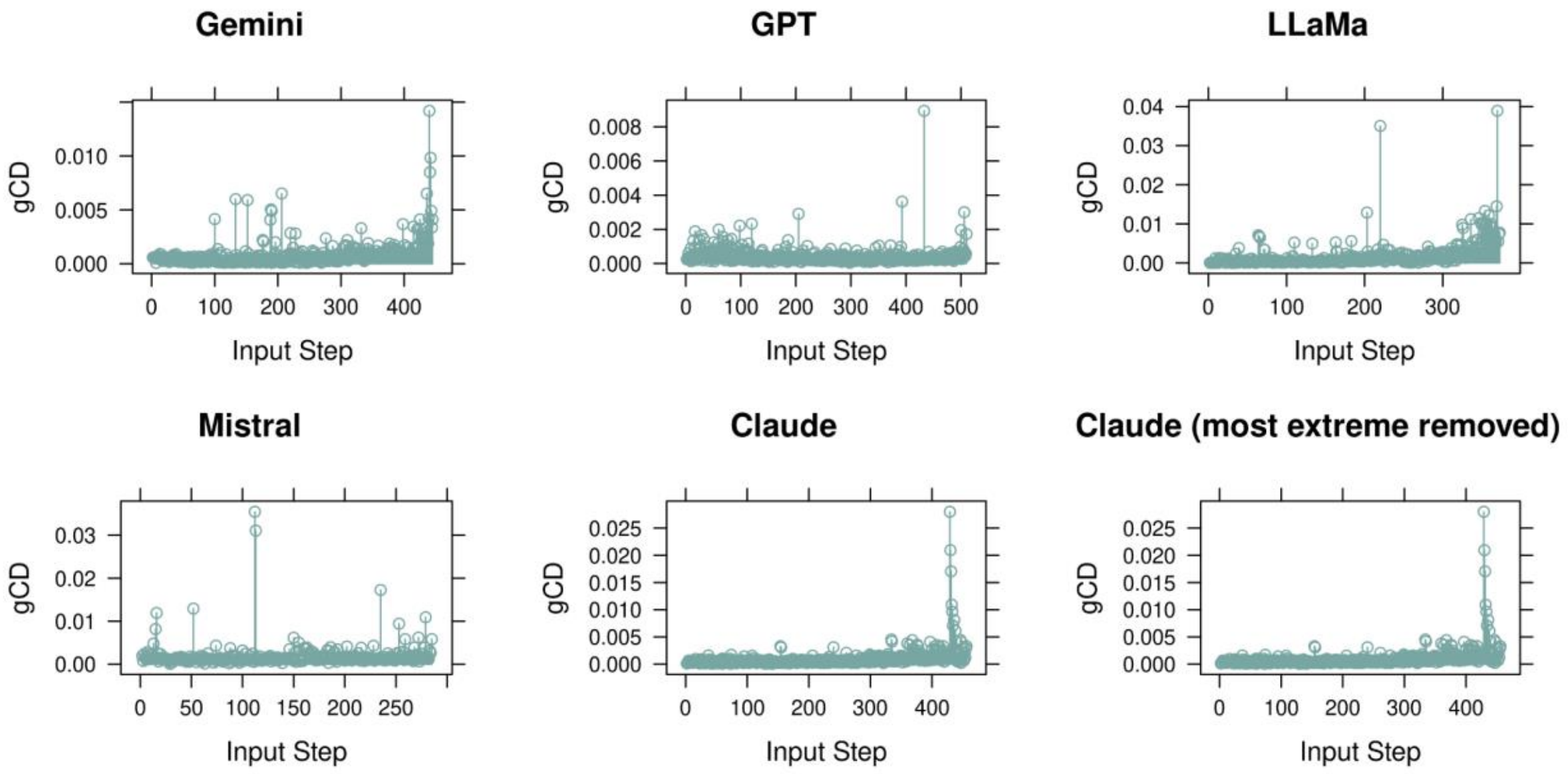

***Figure 6.*** Generalized Cook’s Distances (gCD) by LLM. For Claude, two plots are shown: one includes the most extreme gCD value, while the other excludes it to enhance visual inspection of the remaining observations.

# Discussion

In this study, we assessed LLM-generated responses to the MSLQ in terms of item distributions, the psychological network of SRL constructs, and the psychometric validity of the estimated latent factor structure. In the following sections, we first address the research questions based on the results presented above. We then discuss the limitations and future research directions. Finally, we conclude with a general discussion of the opportunities and threats posed by LLMs for SLR.

## Answers to the Research Questions

> RQ1: *To what extent do item response distributions generated by LLMs exhibit variability in individual item responses and theoretical construct scores?*

Gemini shows notable response variability, including scale-score outliers for most dimensions, whereas GPT, LLaMa, Mistral, and Claude display more stable response patterns across items, with few outliers, if any. This suggests that only one of the five LLMs generates data with substantial sampling variability rather than generating data solely around population means.

> RQ2: *To what extent do LLM-generated datasets reproduce the psychological network structure of the MSLQ, in terms of edge connectivity, strength, directionality, and predictability of SRL processes?*

All models have shown plausible relationships between constructs; however, none of the models provides a complete and coherent picture where all constructs are connected to each other and TA is negatively correlated with either SE or both SE and SR. It was also notable that all the models had high values of explained variance ($R^2$), especially Gemini, which had exceptionally high values with very low errors. These values may be hard to attain in real-life scenarios where data may have large variability, noise, and imperfect responses.

> RQ3: *To what extent do LLM-generated datasets replicate the latent factor structure and inter-construct relationships of the MSLQ, as evidenced by confirmatory and exploratory factor analyses?*

Overall, Gemini (closely followed by Claude) aligns most closely with the 4-factor structure previously identified in empirical research. The main difference is that, in addition to combining CSU and SR, it also merges IV into a broader factor. However, a few items deviate from the pattern, and as a result, new factors emerge. It is possible that this results from LLMs focusing more on linguistic plausibility rather than (only) human psychological authenticity. It is difficult to pinpoint exactly why Gemini and Claude outperform the other models, especially given that both are closed-source and their training procedures, tuning strategies, and error-handling mechanisms are not publicly or clearly disclosed. While one could hypothesize that companies behind closed models invest heavily in infrastructure, alignment, and long-term optimization, this does not fully explain the results, particularly since GPT-4, also a proprietary model, did not perform as well. In saying this, it is important to note that none of the results for the two most promising LLMs seem unreasonable, suggesting that we can potentially learn from these models when developing new questionnaires. They may help reveal how items could be (mis)interpreted compared to how we intend them to be understood.

We conclude that, overall, Gemini was the most promising LLM because it exhibited considerable sampling variability and produced underlying dimensions and theoretical relationships that were closely aligned with established theory and prior empirical work. Although no prior educational research has examined this problem, and existing psychology studies have yielded inconsistent results, our findings demonstrate that LLMs can reproduce plausible factor structures and construct relationships. However, our study also reveals discrepancies and imperfections that highlight both the potential and the current limitations of using LLMs for psychological survey data (e.g., de Winter et al., 2024; López-Pernas et al., 2025).

## Limitations and Future Research

In this study, we did not have a human-completed comparison dataset; therefore, our evaluation of the three research questions relies on theoretically grounded and previous empirical work. This was not a big limitation, given that various studies have previously evaluated the questionnaire's psychometric properties. However, as we move toward persona-based data generation, the lack of theoretical and empirical research on persona-specific human responses makes comparison with human-completed survey data especially important. Additionally, for assessing the plausibility of variability and outliers, a human-completed survey could have been useful.

Relatedly, we have not yet explored generating data for specific personas, based on the reasoning that if LLMs already struggle to produce consistent responses under standard conditions, introducing personas would likely exacerbate these issues. However, we view this as a logical next step; that is, generating data from varied personas and evaluating how well certain LLMs (at minimum, Gemini) produce responses aligned with these profiles. As noted by de Winter et al. (2024), using personas may serve as a valuable initial approach to assess the quality of LLM-generated responses and their potential comprehension of questionnaire items before administering them to human participants. At the same time, Petrov et al. (2024) emphasized that LLMs still face significant challenges in simulating individual-level behavior based on detailed demographic personas. This remains an open question that warrants investigation, specifically in the context of SRL. A challenge in using personas in large questionnaire simulations is the processing load imposed by the length and structure of these. Maintaining a consistent persona across all 44 MSLQ items, for example, requires models to simulate stable beliefs and behaviors, which may exceed the working memory of many LLMs. This is particularly relevant when modelling constructs like SRL, where internal consistency is essential for psychometric validity. Future research could investigate the use of reverse-perspective prompts to potentially minimize some of these challenges. Rather than simulating item-by-item responses, researchers could ask how a specific persona, such as a 22-year-old student with low motivation and test anxiety, would respond to the questionnaire. This global framing may promote more coherent profiles by encouraging the model to reason holistically about the persona rather than defaulting to item-level randomness, although this remains speculative given the opacity of LLM output generation and the limited control over sampling variability.

Moreover, this study does not provide deep insights into the variability in LLM-generated data. In the future, it will be essential to further understand this and, as discussed above, to directly compare it with human-completed surveys. In line with previous findings that LLMs produce more consistent responses with lower variance (e.g., Jansen et al., 2023;  Bisbee et al., 2024), looking at the SDs, we found that only two of the five LLMs under investigation showed considerable variability. As Jansen et al. (2023) noted, this consistency may enhance the reliability of the data. However, if the goal is to simulate real human responses for use in developing interventions or generating recommendations, overly coherent data could present an illusion of validity; that is, the appearance of plausibility without reflecting the true variability and complexity of human behavior.

Another important limitation is that we did not investigate prompt sensitivity, and hence, the extent to which LLM outputs are influenced by the exact phrasing and structure of the input prompts. Even minor changes in wording or parameters can lead to notable differences in responses, which poses challenges for both reproducibility and generalizability. Previous research has demonstrated that the distribution of synthetic responses can be highly sensitive to subtle changes in prompts and may vary considerably over time, even when using the same prompt (Bisbee et al., 2024). Moreover, we did not explore how LLMs respond to longer or more complex questionnaires, which might reveal shifts in style, tone, or coherence across items. Future work should also examine how LLMs handle other formats, such as open-ended questions, continuous rating scales, or mixed-item questionnaires, where variability in logic, consistency, and language style may further affect response quality and interpretability.

## Opportunities and Threats for LLMs in SRL

Beyond the specific limitations and directions for future research, we also deem it essential to critically consider both the general opportunities and threats that LLM-generated data introduce, particularly within the domain of SRL.

### Opportunities

Compared to data collection involving real human participants, the use of LLM-generated data offers a remarkably fast and cost-effective alternative. If the quality of such data proves to be high, it could be very beneficial for preliminary testing. For instance, it allows researchers to examine whether survey items are clearly understood or if specific wording should be revised based on psychometric evaluation, such as assessing whether the intended factor structure aligns with the structure estimated from the generated data. This approach may also serve as a helpful first step in the translation process. Since we generate data from both the original language version and a translated version of a survey, invariance testing via CFA can be used to evaluate equivalence. If misalignments are observed, revisions can be made to the translation before involving human participants for formal validation. Furthermore, model-generated data may better approximate human responses than data simulated under strict population models (López-Pernas et al., 2025), thereby enabling a more realistic evaluation of novel methodologies, such as extensions of network models or predictive tools. One may also investigate whether LLM-generated responses, conditioned on well-constructed contexts or personas (for example), could supplement small human datasets through synthetic augmentation or as a basis for data imputation. While empirical validation on real-world data remains essential, LLM-generated data can serve as an initial testing ground to refine both methodological approaches and survey instruments. This significantly reduces participant burden and minimizes associated costs in the early phases of research development.

Additionally, intervention testing remains a costly exercise with considerable retention problems and limited reach to diverse populations, problems to which simulation with LLM could offer a valuable resource, enabling rapid and low-cost simulations of interventions such as mental health programs or anti-procrastination campaigns. Researchers can test efficacy, optimize content, and identify unintended consequences by prompting LLMs to simulate diverse student personas. A traditional pilot study would require significant funding, face the common issue of students dropping out mid-program (participant retention), and struggle to enroll a truly diverse group representing different majors, backgrounds, and levels of academic distress. LLM simulations could generate plausible student feedback on the intervention and the possible outcomes. For example, no model could successfully replicate the proposed gain when using a program that aims to reduce bullying. Rapidly prototyping and refining intervention models or processes can thus be performed to explore the potential outcomes. Of course, LLMs cannot yet be entirely trusted to simulate the complex, often unpredictable human responses that define genuine learning and psychological change. Their outputs are constrained by patterns in their training data, limiting their ability to meaningfully engage with novel or emotionally nuanced situations. As such, they are best used as tools for early-stage exploration and hypothesis generation, rather than as a substitute for genuine student engagement or empirical validation.

Moreover, the ability to generate survey responses could be potentially a valuable tool for teaching purposes, particularly in areas such as survey design, factor analysis, and statistical methods. For example, as demonstrated by López-Pernas et al. (2025), data generation using LLMs can be effectively integrated into educational activities, which can then be used to apply methodology and statistics. Other studies have demonstrated the utility of LLMs for generating assessment materials, synthetic data for privacy, multimedia, and a wide array of data types (Khalil, Liu, and Jovanovic, 2025).

Finally, generating survey responses using LLMs could also serve as an ethical sandbox, offering a lower-risk environment for prototyping survey instruments on sensitive topics such as mental health, bias, or ethics before involving vulnerable communities. This potential has been explored in previous studies, for instance, to create datasets for suicide prevention (Ghanadian et al., 2024) and to detect hate speech in data-scarce contexts (Khullar et al., 2024). By analyzing LLM-generated outputs for patterns such as biases, dropout tendencies, and demographic distributions, researchers can support the development of intervention tools in areas where data is limited and the focus is on vulnerable populations. Building on this foundation, such practices could contribute to the formulation of ethical guidelines for the use of synthetic data in educational research.

In summary, LLMs can simulate survey responses, test intervention scenarios, explore and refine theoretical models, augment limited datasets, and represent hard-to-reach populations. In essence, they provide a powerful "sandbox" for early-stage educational research, enabling rapid and low-risk experimentation before investing in full-scale human studies.

**Threats**

Unlike humans, LLMs do not generate responses based on internal beliefs, intentions, or self-awareness. Instead, their outputs are constructed from patterns learned during training, based on the statistical likelihood of word sequences. This fundamental difference means that while LLMs can simulate plausible responses, they do not think or self-reflect in the way humans do. As a result, the psychological validity of their responses, especially in constructs that rely on introspection, motivation, or metacognition, remains a key concern.

Another point is that data generated by LLMs is not reproducible, as their outputs are non-deterministic and vary with each generation. This, combined with the ever-changing nature of LLM architectures and parameters, introduces significant challenges for research reproducibility and methodology. Researchers face the challenge of interpreting results when each run produces different outputs, raising concerns about the validity and reliability of synthetic data (Naumova, 2025). Currently, there are no well-established methods to fully address these issues, and the field remains in active development, as reflected by the wide range of methodologies applied to synthetic data generation.

Next, while synthetic data offers practical advantages, especially in low-resource settings, it must be interpreted with caution. Although human-generated data also presents challenges, including potential biases or inaccuracies in self-reporting (Paulhus & Vazire, 2007), it typically allows for transparency throughout the research pipeline, from data collection to analysis and interpretation. In contrast, synthetic data lacks a verifiable connection to real-world behaviour, which raises important questions about how such data should be validated, interpreted, and applied in educational and psychological research. Contextual information can help explain data quality issues or why results may diverge from previous literature or theory. In contrast, LLMs function as black boxes and are often subject to commercial updates beyond the control or awareness of researchers (Ma et al., 2024). These changes can affect the data they generate, raising important questions about how to define the validity of individual data points, such as whether differences in results are due to genuine variation or simply the result of a model update. This lack of transparency poses significant challenges for data validation and interpretation.

Relatedly, assessing the validity of synthetic data in education is challenging because the context is constantly evolving. Student characteristics and skills evolve over time, influenced by personal learning experiences and circumstances, as well as by global events such as the COVID-19 pandemic and emerging technologies like generative AI. These factors can also affect aspects such as students' SRL skills (e.g., Albani et al., 2023; Fan et al., 2025). This challenge has both methodological and ethical implications. Even when synthetic data replicates structural patterns found in previous studies or theory, these insights may not reflect the current students.

Finally, replicating real data presents significant challenges. When synthetic data is used to replicate the characteristics of an average student, it may overlook the diversity and complexity of real student populations. On the other hand, attempts to cover all possible cases regardless of their likelihood can also be problematic (Offenhuber, 2024). For instance, some groups may consist largely of outliers, and relying solely on synthetic data, such as for triangulation with log data, could obscure these critical differences. This may lead to inaccurate insights about specific student groups and result in poorly targeted interventions that do not reflect their actual needs. In some cases, such interventions could even cause harm to students and their learning experiences. This issue is part of a broader challenge within learning analytics, where the misrepresentation of learners through data has long been a concern (Dringus, 2012; Tzimas & Demetriadis, 2021). However, the speed and scale of data generation enabled by LLMs may intensify these risks, especially if their outputs are accepted without critical evaluation (Lissack & Meagher, 2024).

## Appendix A: Data Generation

The prompt used for each generation task was dynamically parameterized to specify the number of students per chunk and the starting student identification number (ID). It included clear instructions to the model to return only CSV-formatted data, with one row per student per item and no additional text:

```
Generate simulated responses for exactly {num_students_this_chunk} unique
students to the 44 MSLQ questions listed below. Start student IDs from
{start_student_id}.

Task:
Create realistic student profiles and their corresponding ratings (1-7)
for ALL 44 MSLQ questions for each student requested. Ensure profiles (name,
background, country, ethnicity, disability, education, age, studies) are
diverse and plausible for educational settings. Maintain consistency within
each student's profile across their 44 question responses.
Output Format:
- Output ONLY the CSV data, starting directly with the first student's
data row.
- DO NOT include the header row in your response.
- DO NOT include any introductory text, explanations, summaries, or
markdown formatting (like ```csv ... ```). Just the raw CSV data lines.
- Each student should have exactly 44 rows, one for each question.
- The columns must be exactly: Student ID, Full_Name, Background,
Country, Ethnicity, Disability, Education Level, Age, Studies, Question
Number, Rating
- Use a comma (,) as the delimiter.
- The 'Rating' column must be an integer between 1 ('not at all true of
me') and 7 ('very true of me').

MSLQ Questions:
{QUESTIONS_TEXT}

Generate the CSV data rows now for {num_students_this_chunk} students
starting with ID {start_student_id}.
```

Due to the substantial volume of output required (each student producing 44 data rows with detailed demographic metadata), the data generation process could not be executed directly via the respective LLM user interfaces (UI), which typically impose stricter output, token, and session limits. We initially attempted this approach using platforms such as ChatGPT and LeChat, both of which returned CSV-formatted output. However, the entire dataset was generated within a few minutes, raising concerns about whether the data were genuinely produced by the underlying model. To further probe data structure and interactions between SRL processes, we conducted a network analysis to examine whether connections emerged among key self-regulated learning dimensions—SR, SE, IV, TA, and CSU. Notably, the UI LLM-generated data revealed no meaningful connections across these dimensions, raising additional concerns about the structural coherence and construct validity of the synthetic responses. To address these limitations, we designed a custom pipeline in Google Colab, integrating API access to each LLM provider and batching the generation process into manageable chunks. “Integrating API access” refers to programmatically connecting with the LLMs via their official application programming interfaces (APIs), which allows direct control over prompt submission, parameter settings (e.g., temperature, max tokens), and structured retrieval of responses, bypassing the limitations of web-based user interfaces. The pipeline followed a modular loop:

1. **Initialize Configuration:** Set model parameters, token limits, and output file paths.
2. **Chunked Request Loop:** Iterate through batches of 10 students per request.
3. **Submit Prompt to API:** Call the selected LLM and capture structured CSV output.
4. **Append to Output File:** Validate and append each batch to a cumulative CSV.
5. **Repeat Until Complete:** Continue until 1,000 students were simulated per model.
6. **Post-validation:** Conduct expert review of data quality, structure, and plausibility.

To monitor performance and ensure reliability, we used both API dashboards (when available) and console-based logging within Colab. For instance, the Google Cloud Console allowed us to visualize request throughput, error rates, and latency during Gemini API calls (see Figure A1). This monitoring revealed that larger batch sizes (e.g., 20-100 students per request) often triggered formatting errors or exceeded token limits, leading to malformed or truncated outputs.

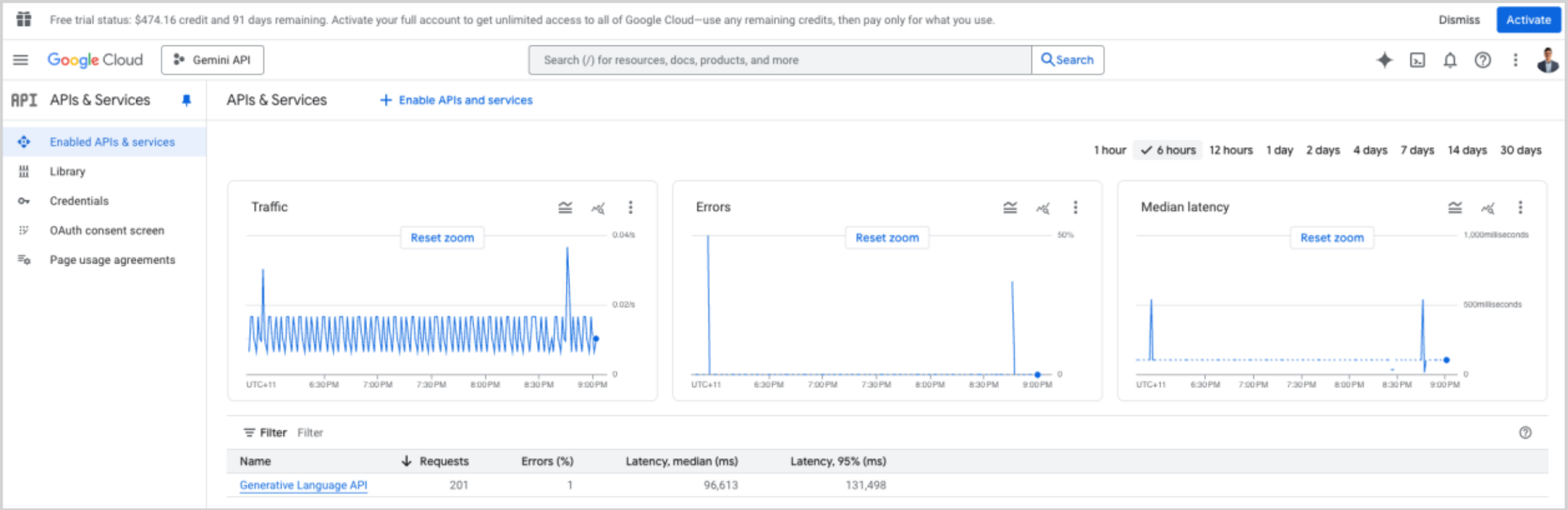

***Figure A1.*** Performance monitoring of Gemini API via Google Cloud Console.

Each batch execution logged the current chunk, the number of simulated students, and the time taken to complete the request. This logging framework was essential for diagnosing API limits (e.g., HTTP 429 errors), managing retries, and maintaining a complete record of the generation process. Smaller batches (10 students per request) ultimately provided the most reliable results, staying within token limits and supporting the model's ability to preserve output formatting. On average, generating 1,000 students per model took approximately 3 hours, with a total runtime of 15 hours across all five LLMs. We preserved the complete set of generated data to facilitate reproducibility and ensure availability for detailed downstream analyses, including validation of inter-construct relationships and model-specific differences in factor structure and SRL network topology.

# Appendix B: Additional Descriptive Statistics

***Table B1***. Descriptive Statistics of LLMs' Responses to Each Questionnaire Item, Grouped by Theoretical Dimensions

| Dimension | Item | Claude | Gemini | GPT | LLaMa | Mistral |
|---|---|---|---|---|---|---|
| CSU | Q23 | M = 5.46 (SD = 1.18) | M = 5.18 (SD = 1.68) | M = 6.00 (SD = 0.71) | M = 5.12 (SD = 1.74) | M = 5.15 (SD = 0.91) |
| | Q24 | M = 5.38 (SD = 0.73) | M = 4.81 (SD = 1.74) | M = 5.90 (SD = 0.78) | M = 5.03 (SD = 1.71) | M = 5.29 (SD = 1.07) |
| | Q26 | M = 3.71 (SD = 1.05) | M = 3.13 (SD = 1.40) | M = 3.46 (SD = 0.82) | M = 4.43 (SD = 1.76) | M = 3.70 (SD = 0.74) |
| | Q28 | M = 4.74 (SD = 0.97) | M = 5.03 (SD = 1.83) | M = 6.06 (SD = 0.71) | M = 5.27 (SD = 1.67) | M = 5.15 (SD = 0.89) |
| | Q29 | M = 5.65 (SD = 0.91) | M = 5.30 (SD = 1.57) | M = 6.13 (SD = 0.77) | M = 5.17 (SD = 1.72) | M = 5.37 (SD = 1.10) |
| | Q30 | M = 5.27 (SD = 0.75) | M = 4.67 (SD = 1.25) | M = 5.57 (SD = 0.73) | M = 4.99 (SD = 1.76) | M = 4.88 (SD = 0.79) |
| | Q31 | M = 4.42 (SD = 0.90) | M = 4.36 (SD = 1.49) | M = 5.29 (SD = 0.82) | M = 4.86 (SD = 1.71) | M = 4.34 (SD = 0.76) |
| | Q34 | M = 4.69 (SD = 0.87) | M = 3.92 (SD = 1.39) | M = 5.51 (SD = 0.76) | M = 5.07 (SD = 1.71) | M = 4.86 (SD = 0.86) |
| | Q36 | M = 5.52 (SD = 1.05) | M = 5.17 (SD = 1.88) | M = 6.15 (SD = 0.76) | M = 4.95 (SD = 1.81) | M = 5.29 (SD = 1.01) |
| | Q39 | M = 5.37 (SD = 1.15) | M = 5.14 (SD = 1.88) | M = 6.21 (SD = 0.75) | M = 5.01 (SD = 1.80) | M = 5.20 (SD = 0.95) |
| | Q41 | M = 4.13 (SD = 1.00) | M = 3.33 (SD = 1.33) | M = 5.04 (SD = 0.80) | M = 5.07 (SD = 1.70) | M = 4.47 (SD = 0.78) |
| | Q42 | M = 4.13 (SD = 1.33) | M = 4.24 (SD = 1.93) | M = 5.39 (SD = 0.83) | M = 4.95 (SD = 1.77) | M = 4.83 (SD = 0.92) |
| | Q44 | M = 5.54 (SD = 1.03) | M = 5.17 (SD = 1.86) | M = 6.17 (SD = 0.73) | M = 5.14 (SD = 1.74) | M = 5.34 (SD = 1.03) |
| IV | Q1 | M = 5.58 (SD = 1.11) | M = 5.52 (SD = 1.62) | M = 5.56 (SD = 0.85) | M = 4.97 (SD = 1.78) | M = 5.51 (SD = 0.85) |
| | Q4 | M = 6.38 (SD = 0.80) | M = 6.06 (SD = 1.41) | M = 6.41 (SD = 0.63) | M = 5.35 (SD = 1.75) | M = 5.63 (SD = 0.86) |
| | Q5 | M = 5.72 (SD = 0.84) | M = 5.49 (SD = 1.51) | M = 5.82 (SD = 0.74) | M = 5.10 (SD = 1.68) | M = 5.78 (SD = 1.00) |
| | Q7 | M = 6.17 (SD = 0.79) | M = 5.64 (SD = 1.67) | M = 6.07 (SD = 0.77) | M = 5.07 (SD = 1.72) | M = 5.11 (SD = 0.88) |
| | Q10 | M = 4.96 (SD = 1.33) | M = 5.31 (SD = 1.95) | M = 5.76 (SD = 0.87) | M = 5.12 (SD = 1.70) | M = 5.44 (SD = 1.18) |
| | Q14 | M = 5.90 (SD = 0.97) | M = 5.66 (SD = 1.54) | M = 6.10 (SD = 0.73) | M = 5.03 (SD = 1.66) | M = 5.42 (SD = 0.94) |
| | Q15 | M = 6.53 (SD = 0.68) | M = 5.97 (SD = 1.57) | M = 6.40 (SD = 0.69) | M = 5.06 (SD = 1.78) | M = 5.75 (SD = 1.02) |
| | Q17 | M = 5.74 (SD = 0.85) | M = 5.45 (SD = 1.68) | M = 6.01 (SD = 0.75) | M = 5.18 (SD = 1.67) | M = 5.58 (SD = 0.95) |
| | Q21 | M = 6.42 (SD = 0.78) | M = 6.00 (SD = 1.53) | M = 6.57 (SD = 0.60) | M = 5.13 (SD = 1.81) | M = 5.81 (SD = 1.00) |
| SE | Q2 | M = 5.22 (SD = 0.95) | M = 4.97 (SD = 1.43) | M = 5.80 (SD = 0.76) | M = 4.98 (SD = 1.63) | M = 4.93 (SD = 0.87) |
| | Q6 | M = 5.19 (SD = 0.94) | M = 5.49 (SD = 1.53) | M = 6.06 (SD = 0.77) | M = 5.02 (SD = 1.78) | M = 5.20 (SD = 0.95) |
| | Q8 | M = 5.23 (SD = 0.95) | M = 4.81 (SD = 1.58) | M = 5.95 (SD = 0.75) | M = 5.12 (SD = 1.76) | M = 5.12 (SD = 0.95) |
| | Q9 | M = 4.79 (SD = 0.96) | M = 4.67 (SD = 1.58) | M = 5.60 (SD = 0.79) | M = 4.91 (SD = 1.72) | M = 4.65 (SD = 0.86) |
| | Q11 | M = 5.14 (SD = 0.95) | M = 5.41 (SD = 1.59) | M = 6.17 (SD = 0.75) | M = 4.89 (SD = 1.71) | M = 5.16 (SD = 0.94) |
| | Q13 | M = 5.20 (SD = 0.96) | M = 5.04 (SD = 1.50) | M = 6.14 (SD = 0.70) | M = 5.08 (SD = 1.78) | M = 4.85 (SD = 0.82) |
| | Q16 | M = 4.30 (SD = 1.07) | M = 4.53 (SD = 1.76) | M = 5.49 (SD = 0.77) | M = 4.96 (SD = 1.68) | M = 4.81 (SD = 0.90) |
| | Q18 | M = 4.13 (SD = 0.97) | M = 4.19 (SD = 1.66) | M = 5.43 (SD = 0.81) | M = 4.98 (SD = 1.72) | M = 4.77 (SD = 0.91) |
| | Q19 | M = 5.55 (SD = 0.88) | M = 5.50 (SD = 1.58) | M = 6.41 (SD = 0.66) | M = 5.10 (SD = 1.82) | M = 5.25 (SD = 0.96) |
| SR | Q25 | M = 4.84 (SD = 1.30) | M = 4.84 (SD = 1.75) | M = 5.84 (SD = 0.81) | M = 5.13 (SD = 1.73) | M = 5.20 (SD = 1.02) |
| | Q27 | M = 2.65 (SD = 0.90) | M = 2.78 (SD = 1.59) | M = 3.22 (SD = 0.78) | M = 4.29 (SD = 1.86) | M = 3.09 (SD = 0.82) |
| | Q32 | M = 4.21 (SD = 1.38) | M = 4.82 (SD = 2.15) | M = 5.73 (SD = 0.86) | M = 5.07 (SD = 1.69) | M = 5.30 (SD = 1.01) |
| | Q33 | M = 5.44 (SD = 1.10) | M = 5.07 (SD = 1.65) | M = 6.06 (SD = 0.77) | M = 5.01 (SD = 1.78) | M = 4.97 (SD = 0.97) |
| | Q35 | M = 4.92 (SD = 1.29) | M = 4.69 (SD = 2.03) | M = 6.07 (SD = 0.76) | M = 5.02 (SD = 1.72) | M = 4.90 (SD = 0.96) |
| | Q37 | M = 3.45 (SD = 1.06) | M = 3.21 (SD = 1.78) | M = 3.43 (SD = 0.73) | M = 4.66 (SD = 1.74) | M = 3.40 (SD = 0.79) |
| | Q38 | M = 2.83 (SD = 1.08) | M = 2.95 (SD = 1.62) | M = 3.45 (SD = 0.77) | M = 4.62 (SD = 1.80) | M = 3.43 (SD = 0.87) |
| | Q40 | M = 5.06 (SD = 1.10) | M = 4.88 (SD = 1.84) | M = 5.68 (SD = 0.74) | M = 5.17 (SD = 1.72) | M = 5.09 (SD = 1.04) |
| | Q43 | M = 6.07 (SD = 0.83) | M = 5.58 (SD = 1.66) | M = 6.47 (SD = 0.66) | M = 5.08 (SD = 1.76) | M = 5.18 (SD = 0.98) |
| TA | Q3 | M = 4.68 (SD = 1.15) | M = 3.47 (SD = 1.61) | M = 3.82 (SD = 0.85) | M = 3.97 (SD = 1.76) | M = 3.39 (SD = 1.00) |
| | Q12 | M = 4.73 (SD = 1.10) | M = 3.68 (SD = 1.97) | M = 3.73 (SD = 0.77) | M = 4.18 (SD = 1.81) | M = 3.87 (SD = 0.86) |
| | Q20 | M = 4.56 (SD = 1.16) | M = 3.51 (SD = 2.05) | M = 3.71 (SD = 0.77) | M = 4.69 (SD = 1.72) | M = 4.00 (SD = 0.81) |
| | Q22 | M = 3.93 (SD = 1.07) | M = 2.82 (SD = 1.85) | M = 3.52 (SD = 0.80) | M = 4.47 (SD = 1.76) | M = 3.40 (SD = 0.83) |

*Note*. Items 26, 27, 37, and 38 have their original coding (not reversed).

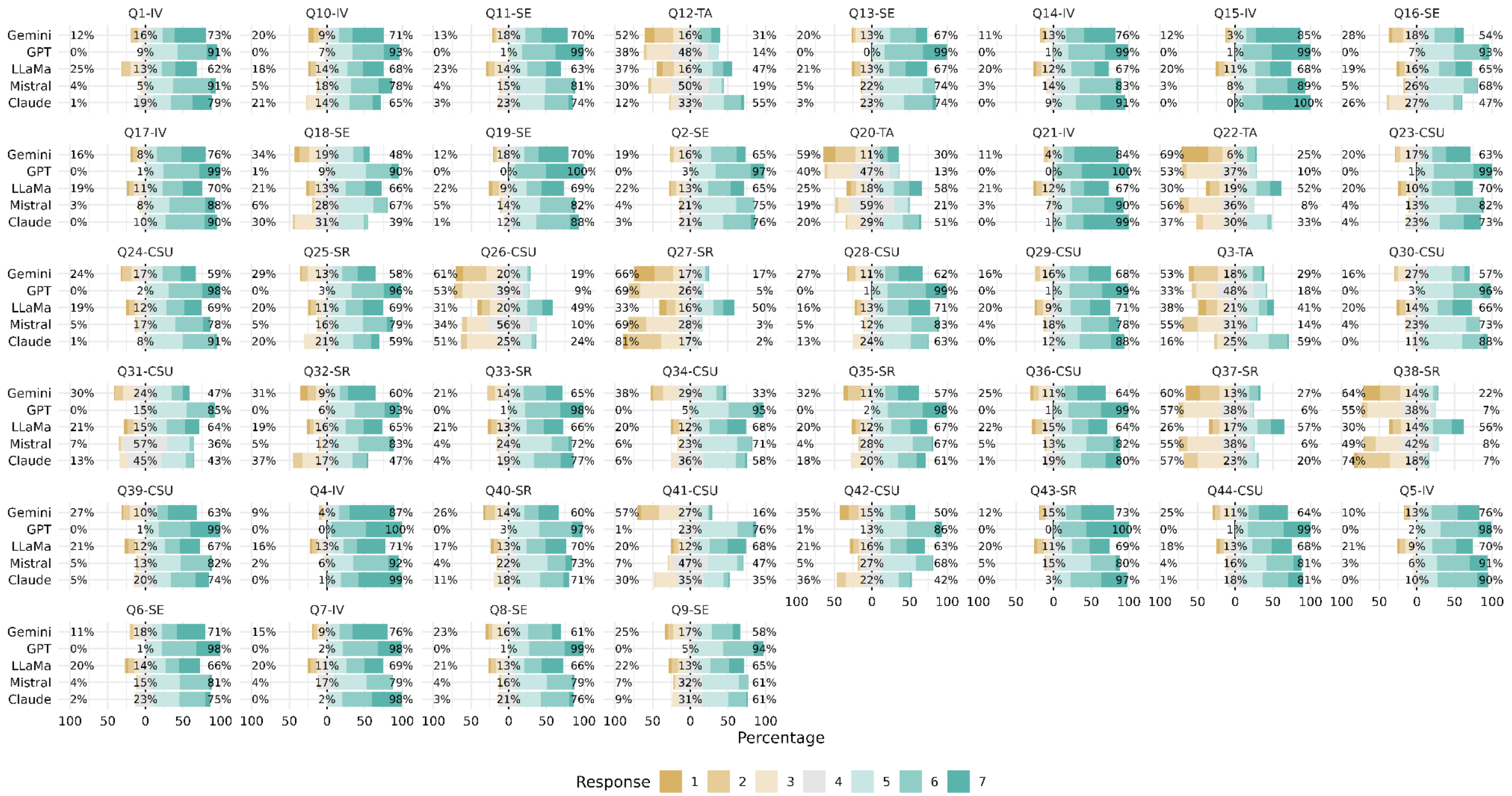

***Figure B1.*** Response distributions generated by the five LLMs. Each horizontal bar represents the frequency of responses on a Likert-type scale. Turquoise/greenish bars indicate higher ratings, while beige/orange bars reflect lower ratings. The gray vertical line marks the scale midpoint (3). Percentages on the left and right of each item denote the proportion of responses below and above the midpoint, respectively. Note that items 26, 27, 37, and 38 have their original coding (not reversed).

# Appendix C: Missing Data

In addition to the main analyses, we explored missing data patterns to understand which missing data mechanism is underlying the data generated by the LLMs. This is relevant to the network and factor analyses in this study, as well as for any future studies that utilize the item responses. In general, analyses can accommodate missing data (when estimated using full information maximum likelihood estimation or using multiple imputation), when the data are at least missing at random (MAR); that is, there is a reason for missingness, which is also related to studied variables, but the analysis can account for this by including the reason as an observed variable in the analysis. If the missing data are entirely randomly scattered throughout the responses without any systematic cause, the data are considered missing completely at random (MCAR), which is even simpler to account for than MAR because even a complete-case analysis would lead to unbiased results. In LLM-generated data, it is most plausible that missingness is related to item position as a consequence of the LLMs' memory or token constraints during generation. However, since the item position is unrelated to the actual item responses, the missing data mechanism can be regarded as functionally equivalent to MCAR. To examine this further, we inspect the patterns of missing values across each LLM dataset using the R package *mice* (van Buuren & Groothuis-Oudshoorn, 2011).

The percentage of missing data per model was as follows: Gemini had 16.21% missing, GPT 13.92%, LLaMa 35.39%, Mistral 22.08%, and Claude3 14.78%. Figure C2 shows the missing data patterns. Columns represent variables (ordered by item number and thus sequence in which they were completed), and rows correspond to unique patterns of missingness. Blue cells indicate observed values, while orange cells represent missing data. For Gemini, GPT, Mistral, and Claude, a distinct staircase pattern is evident, indicating that missing values, if they occur for one item, also occur in variables recorded at later time points. Further inspection revealed that data generation typically processed correctly up to a certain student in a given batch, for whom only a partial set of responses (e.g., 34 out of 44) was produced before abruptly stopping. After this cutoff, no responses were generated for any of the remaining individuals in that batch. Generation then resumed with the first student in the next batch of ten, starting from the explicitly defined ID. These interruptions appeared to be tied to internal LLM limitations, such as token exhaustion or breakdowns in formatting and state continuity. In these cases, our batching approach—where each new batch begins from an explicitly defined starting ID—ensured that subsequent data generation resumed correctly and independently of earlier failures. As a result, the dataset was ultimately compiled in full through repeated, controlled iterations, even when individual batches were affected by errors.

Notably, only in Mistral, for ten patterns, observed data reappear after missing data; that is, data appear, disappear, and then reappear. Note that each of these patterns occurs in only a single observation, and all but the first occur in a sequence, which is not visible in the figure but has been verified through additional exploration. Although it is unclear what has happened here, the issue affects only 1.6 percent of the observations, which we do not deem concerning; the records were not removed from further analyses. Importantly, we conclude that the missingness in all datasets can be classified as MCAR, as these missing data were attributable to internal model-level issues rather than any systematic variation in item content or student characteristics.

Investigation of the missing data patterns for LLaMa indicates they are entirely unstructured. The missingness appears random and unrelated to specific item or dimension combinations, supporting the assumption of MCAR. Nonetheless, future research should examine the underlying causes of this pattern.

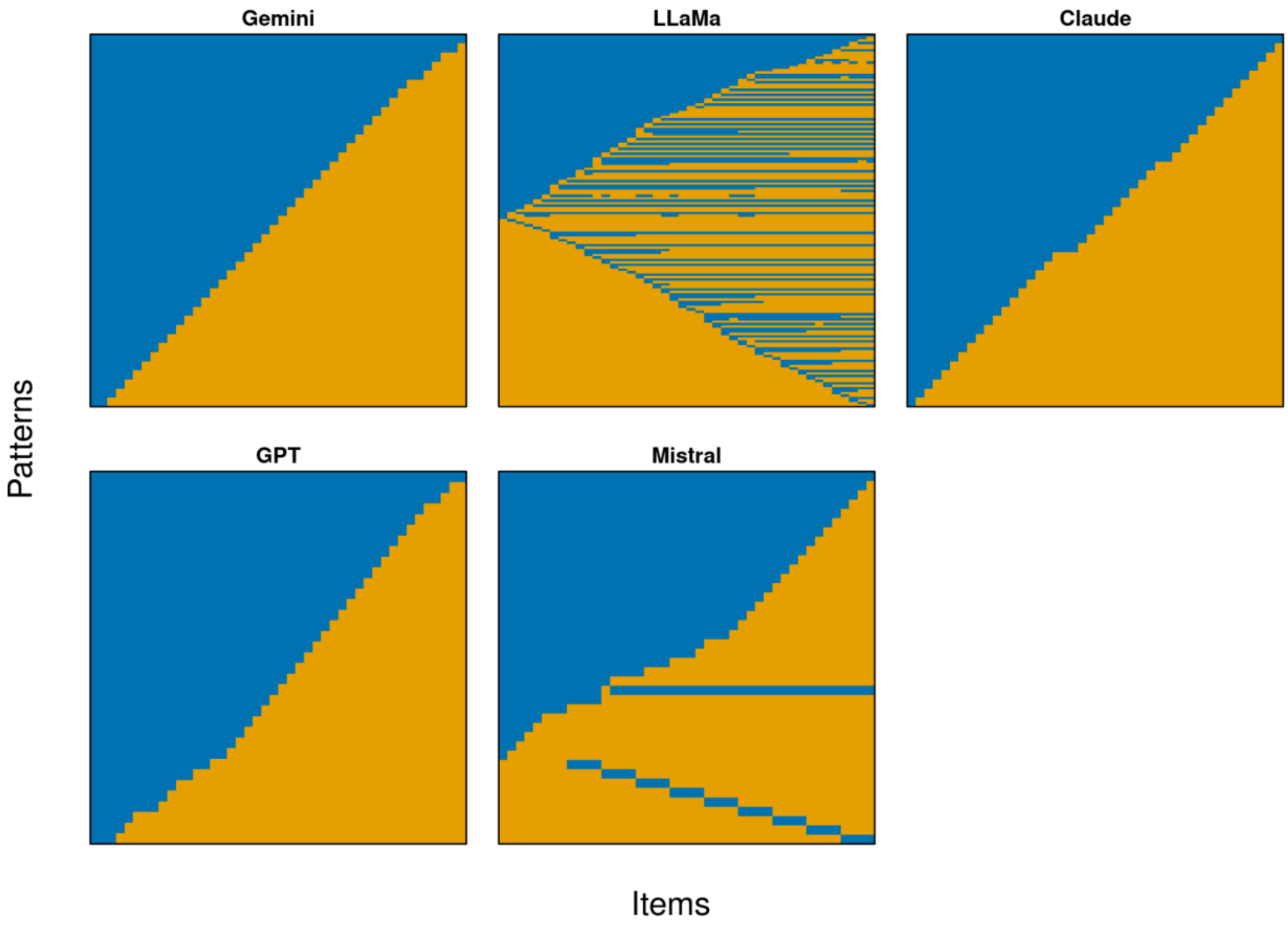

***Figure C2.*** Missing data patterns for the four LLMs with missing data. Columns represent variables (in item order), rows show unique missingness patterns. Blue = observed, orange = missing.

***Table C1.*** Number of Missing Items for Each LLM per Questionnaire Item

| Dimension | Item | Gemini | GPT | LLaMa | Mistral | Claude |
|---|---|---|---|---|---|---|
| IV | Q1 | 0 | 0 | 127 | 10 | 0 |
| | Q4 | 7 | 1 | 134 | 15 | 13 |
| | Q5 | 8 | 2 | 136 | 16 | 19 |
| | Q7 | 16 | 6 | 143 | 17 | 24 |
| | Q10 | 27 | 9 | 161 | 17 | 31 |
| | Q14 | 38 | 12 | 205 | 19 | 47 |
| | Q15 | 39 | 14 | 210 | 19 | 51 |
| | Q17 | 51 | 15 | 219 | 19 | 57 |
| | Q21 | 69 | 23 | 230 | 21 | 65 |
| SE | Q2 | 0 | 0 | 133 | 10 | 3 |
| | Q6 | 14 | 6 | 137 | 17 | 20 |
| | Q8 | 19 | 6 | 147 | 17 | 25 |
| | Q9 | 24 | 7 | 149 | 17 | 28 |
| | Q11 | 33 | 10 | 190 | 17 | 33 |
| | Q13 | 37 | 12 | 199 | 19 | 45 |
| | Q16 | 44 | 14 | 214 | 19 | 56 |
| | Q18 | 57 | 16 | 222 | 20 | 62 |
| | Q19 | 63 | 18 | 222 | 20 | 62 |
| TA | Q3 | 3 | 0 | 133 | 11 | 5 |
| | Q12 | 35 | 10 | 199 | 17 | 37 |
| | Q20 | 67 | 21 | 224 | 20 | 62 |
| | Q22 | 71 | 27 | 234 | 21 | 68 |
| CSU | Q23 | 74 | 32 | 241 | 21 | 70 |
| | Q24 | 80 | 40 | 239 | 24 | 73 |
| | Q26 | 82 | 49 | 241 | 26 | 82 |
| | Q28 | 89 | 55 | 247 | 28 | 87 |
| | Q29 | 91 | 60 | 253 | 32 | 91 |
| | Q30 | 95 | 71 | 250 | 35 | 94 |
| | Q31 | 98 | 89 | 256 | 39 | 94 |
| | Q34 | 104 | 108 | 257 | 56 | 100 |
| | Q36 | 112 | 116 | 259 | 78 | 107 |
| | Q39 | 125 | 124 | 266 | 112 | 117 |
| | Q41 | 133 | 130 | 269 | 120 | 122 |
| | Q42 | 138 | 136 | 270 | 123 | 126 |
| | Q44 | 143 | 137 | 272 | 132 | 132 |
| SR | Q25 | 81 | 42 | 241 | 26 | 77 |
| | Q27 | 86 | 53 | 247 | 26 | 84 |
| | Q32 | 101 | 94 | 256 | 44 | 96 |
| | Q33 | 103 | 102 | 257 | 48 | 99 |
| | Q35 | 110 | 113 | 258 | 65 | 104 |
| | Q37 | 119 | 120 | 261 | 91 | 111 |
| | Q38 | 125 | 122 | 264 | 103 | 113 |
| | Q40 | 126 | 130 | 266 | 116 | 119 |
| | Q43 | 138 | 137 | 272 | 124 | 131 |

**References Appendix C**

# Appendix D: Factor Loadings EFA

***Table D1.*** EFA factor Loadings per LLM, Sorted by Theoretical Dimensions

| | | Gemini | | | | LLaMa | | | | | | | Mistral | | | | | GPT | | | | | | Claude | | | | |
|---|---|---|---|---|---|---|---|---|---|---|---|---|---|---|---|---|---|---|---|---|---|---|---|---|---|---|---|---|
| **Dim.** | **Item** | **f1** | **f2** | **f3** | **f4** | **f1** | **f2** | **f3** | **f4** | **f5** | **f6** | **f7** | **f1** | **f2** | **f3** | **f4** | **f5** | **f1** | **f2** | **f3** | **f4** | **f5** | **f6** | **f1** | **f2** | **f3** | **f4** | **f5** |
| | **Q23** | 0.12 | 0.02 | 0.88 | 0.03 | −0.01 | 0.02 | 0.69 | 0.20 | −0.02 | −0.14 | 0.03 | −0.06 | −0.01 | 0.02 | 0.95 | −0.21 | −0.03 | 0.03 | −0.06 | −0.07 | 0.79 | 0.02 | 0.06 | −0.07 | −0.01 | 0.01 | 0.91 |
| | **Q24** | 0.01 | −0.01 | 0.95 | 0.15 | −0.05 | 0.81 | 0.02 | 0.02 | −0.02 | 0.33 | −0.02 | 0.01 | 0.04 | −0.01 | 0.91 | 0.29 | 0.00 | 0.14 | −0.02 | 0.06 | 0.71 | 0.06 | 0.20 | −0.13 | −0.01 | 0.27 | 0.60 |
| | **Q26** | −0.58 | 0.07 | −0.41 | −0.04 | 0.55 | −0.01 | 0.77 | 0.00 | 0.22 | 0.08 | −0.01 | 0.06 | 0.83 | 0.08 | 0.04 | 0.00 | 0.02 | 0.00 | 0.61 | 0.43 | 0.00 | −0.09 | −0.27 | 0.07 | −0.02 | 0.41 | −0.43 |
| | **Q28** | 0.10 | −0.04 | 0.90 | −0.02 | −0.14 | 0.00 | −0.07 | 0.90 | 0.04 | 0.24 | 0.00 | −0.06 | −0.01 | −0.01 | 0.96 | −0.20 | −0.05 | −0.01 | −0.05 | −0.01 | 0.78 | 0.11 | −0.04 | 0.12 | 0.16 | −0.04 | 0.97 |
| | **Q29** | 0.03 | 0.11 | 0.92 | 0.02 | −0.04 | 0.01 | 0.95 | −0.02 | −0.10 | −0.01 | 0.24 | 0.00 | −0.02 | 0.01 | 0.87 | 0.32 | 0.03 | −0.01 | −0.01 | 0.17 | 0.78 | −0.04 | 0.06 | 0.12 | −0.06 | 0.03 | 0.87 |
| | **Q30** | 0.12 | 0.07 | 0.67 | 0.54 | 0.00 | 0.86 | −0.07 | 0.02 | 0.37 | −0.01 | −0.03 | 0.04 | 0.02 | −0.08 | 0.84 | −0.22 | 0.10 | 0.01 | −0.02 | 0.04 | 0.58 | 0.29 | 0.37 | −0.32 | −0.03 | 0.58 | 0.10 |
| **CSU** | **Q31** | −0.06 | −0.04 | 0.91 | 0.34 | 0.23 | 0.01 | 0.00 | 0.89 | −0.01 | −0.23 | 0.00 | 0.04 | 0.17 | −0.38 | 0.78 | 0.01 | 0.05 | −0.05 | −0.03 | 0.25 | 0.56 | 0.36 | 0.00 | −0.23 | 0.04 | 0.65 | 0.14 |
| | **Q34** | 0.00 | 0.00 | 0.74 | 0.60 | 0.04 | 0.07 | 0.01 | 0.74 | 0.25 | 0.01 | 0.00 | 0.05 | 0.09 | −0.13 | 0.88 | −0.20 | −0.01 | −0.03 | −0.03 | −0.02 | 0.54 | 0.42 | 0.26 | −0.26 | 0.03 | 0.71 | 0.00 |
| | **Q36** | 0.09 | 0.03 | 0.91 | −0.04 | 0.03 | 0.73 | 0.08 | 0.02 | −0.01 | 0.37 | 0.03 | −0.03 | 0.02 | −0.03 | 0.93 | 0.09 | 0.11 | 0.02 | −0.06 | 0.03 | 0.71 | −0.08 | 0.17 | −0.07 | −0.04 | −0.10 | 0.76 |
| | **Q39** | 0.12 | −0.01 | 0.89 | −0.04 | 0.01 | 0.85 | 0.08 | 0.01 | −0.14 | −0.10 | −0.12 | −0.02 | −0.06 | 0.01 | 0.92 | −0.19 | −0.03 | −0.16 | −0.03 | 0.05 | 0.86 | −0.01 | 0.09 | 0.12 | 0.06 | −0.03 | 0.91 |
| | **Q41** | −0.04 | −0.02 | 0.74 | 0.57 | 0.07 | −0.02 | 0.93 | 0.00 | 0.02 | −0.03 | 0.31 | −0.01 | 0.19 | −0.25 | 0.83 | −0.14 | −0.02 | −0.13 | 0.05 | 0.16 | 0.57 | 0.41 | 0.01 | −0.18 | 0.10 | 0.68 | −0.06 |
| | **Q42** | 0.03 | −0.11 | 0.94 | 0.14 | 0.05 | 0.87 | 0.01 | −0.07 | 0.38 | −0.01 | 0.01 | −0.02 | −0.03 | −0.23 | 0.90 | 0.09 | −0.01 | 0.05 | 0.00 | 0.32 | 0.61 | 0.29 | 0.00 | −0.08 | 0.17 | 0.17 | 0.93 |
| | **Q44** | 0.11 | 0.02 | 0.89 | −0.05 | 0.00 | 0.02 | 0.75 | 0.08 | −0.12 | 0.40 | −0.01 | 0.04 | 0.00 | 0.03 | 0.86 | 0.11 | −0.04 | −0.09 | 0.00 | −0.05 | 0.80 | 0.01 | 0.02 | 0.18 | 0.09 | −0.05 | 0.92 |
| | **Q1** | −0.01 | 0.22 | 0.91 | 0.00 | −0.14 | 0.01 | 0.37 | 0.35 | 0.04 | −0.01 | 0.03 | −0.04 | −0.13 | 0.17 | 0.80 | −0.03 | 0.05 | 0.01 | −0.10 | −0.07 | 0.68 | 0.04 | 0.06 | 0.02 | −0.20 | 0.07 | 0.83 |
| | **Q4** | −0.14 | 0.49 | 0.92 | 0.04 | −0.16 | 0.00 | 0.03 | 0.78 | 0.02 | 0.12 | 0.06 | −0.05 | −0.09 | 0.22 | 0.79 | −0.05 | −0.07 | −0.09 | 0.00 | −0.02 | 0.79 | −0.22 | −0.12 | 0.01 | −0.45 | 0.01 | 0.83 |
| | **Q5** | 0.18 | 0.27 | 0.64 | −0.06 | −0.04 | 0.07 | 0.75 | 0.07 | 0.02 | 0.01 | 0.16 | 0.06 | 0.16 | 0.21 | 0.84 | −0.02 | −0.22 | 0.34 | 0.01 | −0.03 | 0.73 | 0.01 | 0.02 | 0.77 | −0.03 | 0.02 | 0.73 |
| | **Q7** | 0.17 | 0.31 | 0.76 | −0.03 | −0.04 | 0.08 | 0.00 | 0.79 | 0.06 | −0.24 | −0.01 | 0.06 | 0.05 | −0.08 | 0.77 | 0.02 | −0.03 | −0.15 | 0.03 | 0.00 | 0.66 | −0.07 | 0.22 | 0.00 | −0.32 | −0.10 | 0.58 |
| **IV** | **Q10** | −0.02 | 0.22 | 0.94 | −0.08 | −0.04 | −0.01 | 0.02 | 0.83 | 0.21 | 0.00 | 0.00 | −0.19 | −0.09 | 0.23 | 0.85 | 0.04 | −0.18 | 0.00 | −0.03 | −0.03 | 0.60 | 0.18 | −0.18 | 0.16 | −0.10 | 0.01 | 0.96 |
| | **Q14** | −0.09 | 0.25 | 0.98 | −0.02 | 0.04 | 0.01 | 0.86 | −0.02 | 0.14 | 0.10 | 0.00 | −0.02 | 0.02 | 0.08 | 0.89 | 0.07 | −0.09 | 0.03 | −0.03 | −0.04 | 0.67 | 0.00 | −0.09 | 0.07 | −0.17 | −0.06 | 0.89 |
| | **Q15** | 0.01 | 0.45 | 0.83 | −0.05 | −0.03 | 0.80 | 0.05 | 0.06 | 0.00 | −0.15 | −0.05 | 0.01 | 0.09 | 0.32 | 0.86 | −0.09 | 0.02 | 0.00 | 0.00 | −0.02 | 0.75 | −0.20 | 0.00 | 0.27 | −0.39 | −0.02 | 0.73 |
| | **Q17** | 0.19 | 0.28 | 0.60 | −0.11 | −0.01 | 0.00 | 0.87 | 0.04 | 0.03 | −0.04 | 0.27 | 0.01 | 0.02 | 0.17 | 0.89 | 0.08 | −0.22 | 0.13 | 0.02 | 0.02 | 0.83 | −0.01 | −0.02 | 0.79 | 0.00 | 0.01 | 0.73 |
| | **Q21** | −0.01 | 0.50 | 0.84 | 0.00 | −0.08 | 0.78 | 0.16 | −0.03 | 0.01 | −0.02 | 0.22 | 0.02 | 0.07 | 0.30 | 0.90 | 0.00 | −0.02 | −0.01 | −0.02 | −0.07 | 0.82 | −0.28 | −0.10 | 0.12 | −0.44 | 0.01 | 0.82 |
| | **Q2** | 0.97 | 0.01 | −0.01 | 0.17 | −0.03 | 0.09 | 0.55 | 0.18 | 0.12 | 0.02 | 0.02 | 0.59 | −0.13 | 0.00 | 0.46 | 0.01 | 0.54 | 0.10 | −0.01 | 0.00 | 0.46 | 0.03 | 1.06 | −0.01 | 0.01 | 0.03 | −0.13 |
| | **Q6** | 1.00 | 0.02 | −0.01 | 0.00 | −0.04 | 0.68 | 0.16 | 0.00 | 0.14 | 0.05 | 0.04 | 0.19 | −0.06 | 0.14 | 0.78 | 0.01 | 0.37 | −0.17 | −0.05 | −0.08 | 0.60 | 0.00 | 0.91 | 0.02 | 0.01 | −0.05 | 0.04 |
| | **Q8** | 0.82 | 0.06 | 0.20 | 0.12 | −0.03 | 0.02 | 0.71 | 0.14 | −0.03 | 0.21 | 0.02 | 0.35 | −0.03 | 0.12 | 0.74 | 0.00 | 0.42 | 0.25 | −0.02 | 0.00 | 0.54 | −0.01 | 1.07 | 0.00 | 0.00 | 0.02 | −0.13 |
| | **Q9** | 0.97 | −0.04 | −0.01 | 0.14 | 0.06 | 0.72 | 0.24 | −0.04 | 0.00 | 0.03 | 0.20 | 0.50 | 0.02 | −0.11 | 0.60 | −0.06 | 0.32 | 0.02 | 0.06 | −0.05 | 0.58 | 0.15 | 0.84 | 0.14 | 0.28 | 0.00 | −0.01 |
| **SE** | **Q11** | 0.91 | 0.06 | 0.10 | 0.01 | 0.12 | 0.00 | 0.81 | 0.08 | −0.04 | −0.12 | −0.01 | 0.27 | −0.04 | 0.01 | 0.81 | −0.04 | 0.35 | −0.13 | 0.01 | 0.01 | 0.66 | −0.02 | 0.87 | 0.00 | 0.00 | −0.03 | 0.11 |
| | **Q13** | 0.77 | 0.07 | 0.24 | 0.11 | 0.04 | −0.01 | 0.00 | 0.92 | −0.11 | 0.03 | 0.33 | 0.30 | −0.01 | 0.04 | 0.75 | 0.03 | 0.27 | 0.12 | 0.00 | 0.01 | 0.68 | −0.04 | 1.01 | 0.00 | 0.01 | −0.01 | −0.08 |
| | **Q16** | 0.77 | −0.14 | 0.21 | 0.14 | 0.06 | 0.03 | 0.02 | 0.79 | −0.01 | 0.21 | −0.01 | 0.25 | −0.06 | −0.12 | 0.79 | 0.08 | 0.25 | 0.07 | 0.02 | 0.00 | 0.58 | 0.27 | 0.79 | 0.00 | 0.16 | −0.03 | 0.16 |
| | **Q18** | 1.02 | −0.14 | −0.15 | 0.17 | −0.03 | 0.85 | −0.06 | 0.03 | 0.29 | 0.03 | −0.02 | 0.22 | 0.01 | −0.10 | 0.84 | 0.11 | 0.30 | −0.03 | 0.04 | −0.01 | 0.57 | 0.25 | 0.78 | 0.11 | 0.29 | 0.01 | 0.08 |
| | **Q19** | 0.92 | 0.06 | 0.09 | −0.01 | −0.06 | −0.01 | 0.07 | 0.84 | −0.08 | −0.12 | −0.01 | 0.20 | −0.02 | 0.10 | 0.83 | −0.05 | 0.18 | −0.12 | 0.02 | 0.02 | 0.80 | −0.19 | 0.75 | 0.01 | −0.06 | −0.09 | 0.17 |
| | **Q25** | 0.18 | −0.02 | 0.82 | 0.05 | −0.01 | 0.02 | 0.04 | 0.85 | 0.02 | −0.07 | 0.26 | 0.00 | −0.06 | −0.02 | 0.84 | −0.08 | 0.05 | −0.12 | 0.05 | −0.03 | 0.67 | 0.18 | −0.05 | 0.07 | 0.15 | −0.01 | 0.99 |
| | **Q27** | −0.21 | 0.00 | −0.77 | 0.00 | 0.70 | 0.77 | 0.02 | 0.00 | 0.01 | −0.14 | 0.02 | −0.01 | 0.60 | −0.13 | −0.24 | −0.02 | −0.01 | 0.06 | 0.54 | 0.22 | −0.10 | 0.01 | −0.16 | 0.04 | −0.01 | 0.37 | −0.62 |
| | **Q32** | 0.02 | −0.02 | 0.97 | −0.01 | 0.01 | 0.01 | 0.79 | 0.03 | 0.01 | 0.36 | 0.00 | −0.02 | −0.07 | 0.01 | 0.89 | 0.01 | 0.03 | 0.04 | −0.02 | 0.14 | 0.64 | 0.19 | 0.03 | 0.05 | 0.09 | 0.03 | 0.95 |
| | **Q33** | 0.02 | 0.10 | 0.93 | 0.08 | −0.01 | 0.83 | −0.01 | 0.08 | −0.06 | 0.01 | 0.20 | 0.00 | −0.02 | −0.12 | 0.87 | 0.13 | 0.01 | 0.22 | 0.04 | 0.14 | 0.76 | −0.06 | 0.08 | 0.01 | 0.00 | −0.06 | 0.85 |
| **SR** | **Q35** | 0.05 | −0.06 | 0.94 | 0.01 | 0.03 | −0.02 | 0.82 | 0.10 | −0.02 | −0.19 | −0.04 | −0.01 | −0.12 | −0.08 | 0.80 | −0.02 | 0.02 | −0.16 | 0.00 | 0.08 | 0.78 | 0.02 | 0.00 | 0.01 | 0.14 | −0.05 | 0.94 |
| | **Q37** | −0.36 | 0.10 | −0.65 | 0.00 | 0.53 | −0.01 | −0.03 | 0.90 | −0.02 | 0.00 | 0.44 | 0.02 | 0.78 | −0.10 | −0.10 | −0.08 | −0.06 | −0.06 | 0.66 | 0.35 | −0.07 | 0.00 | −0.16 | 0.03 | −0.04 | 0.39 | −0.59 |
| | **Q38** | −0.10 | 0.07 | −0.85 | −0.01 | 0.58 | 0.04 | 0.66 | 0.02 | 0.29 | 0.00 | −0.01 | −0.05 | 0.68 | 0.00 | −0.09 | 0.07 | −0.03 | 0.05 | 0.66 | 0.17 | 0.01 | 0.01 | −0.04 | 0.04 | 0.00 | 0.33 | −0.71 |
| | **Q40** | −0.01 | −0.04 | 0.98 | 0.09 | −0.13 | −0.02 | 0.00 | 0.83 | 0.03 | 0.27 | −0.02 | −0.01 | 0.02 | −0.15 | 0.89 | 0.34 | −0.04 | 0.01 | 0.01 | 0.10 | 0.74 | 0.22 | −0.21 | −0.04 | 0.16 | 0.09 | 1.00 |
| | **Q43** | −0.09 | 0.28 | 0.98 | 0.02 | 0.02 | 0.06 | −0.16 | 0.99 | −0.06 | −0.21 | −0.02 | −0.02 | 0.00 | 0.05 | 0.88 | −0.04 | 0.02 | −0.01 | −0.01 | 0.12 | 0.86 | −0.15 | 0.34 | −0.08 | −0.23 | 0.00 | 0.54 |
| | **Q3** | −0.87 | 0.21 | 0.02 | 0.05 | 0.60 | 0.70 | −0.09 | 0.06 | −0.02 | −0.02 | 0.00 | −0.12 | 0.73 | −0.05 | −0.20 | −0.01 | 0.01 | 0.05 | 0.70 | −0.03 | 0.06 | 0.03 | −0.04 | 0.02 | 0.16 | 0.89 | −0.07 |
| **TA** | **Q12** | −0.83 | 0.23 | −0.06 | 0.04 | 0.62 | 0.81 | −0.13 | −0.01 | 0.09 | 0.24 | 0.00 | −0.12 | 0.91 | 0.17 | −0.03 | 0.02 | 0.09 | −0.02 | 0.77 | −0.13 | 0.00 | −0.02 | −0.04 | 0.01 | 0.06 | 0.95 | −0.01 |
| | **Q20** | −0.84 | 0.21 | −0.05 | 0.05 | 0.42 | −0.02 | 0.81 | −0.02 | 0.00 | 0.29 | −0.03 | −0.12 | 0.90 | 0.26 | 0.02 | 0.03 | −0.01 | 0.01 | 0.72 | 0.02 | 0.01 | 0.01 | −0.08 | −0.06 | −0.02 | 0.90 | −0.01 |
| | **Q22** | −0.76 | 0.18 | −0.16 | 0.11 | 0.56 | −0.01 | 0.02 | 0.71 | 0.32 | 0.00 | 0.04 | −0.12 | 0.81 | 0.01 | −0.09 | −0.02 | 0.08 | 0.00 | 0.76 | −0.01 | 0.01 | 0.12 | 0.01 | −0.02 | 0.08 | 0.89 | −0.18 |

*Note.* Items 26, 27, 37, and 38 have their original coding (not reversed).